\documentclass{article}
\pdfoutput=1

\usepackage[utf8]{inputenc} 
\usepackage[T1]{fontenc}    
\PassOptionsToPackage{hyphens}{url} 
\usepackage{hyperref}       
\usepackage{booktabs}       
\usepackage{amsfonts}       
\usepackage{nicefrac}       
\usepackage{microtype}      
\usepackage{xcolor}         
\usepackage{algorithm}
\usepackage{algorithmic}
\usepackage{graphicx}
\usepackage{enumerate}
\usepackage{subfig}
\usepackage{comment}
\usepackage[export]{adjustbox}
\usepackage{mathtools}

\usepackage{authblk}
\usepackage{fullpage}

\usepackage{enumitem}

\usepackage{natbib}

\usepackage{Defs}

\title{Designing Decision Support Systems Using \\ Counterfactual Prediction Sets}

\author{Eleni Straitouri}
\author{Manuel Gomez Rodriguez}

\affil{Max Planck Institute for Software Systems \\ \{estraitouri, manuel\}@mpi-sws.org}

\date{}

\begin{document}
\maketitle

\begin{abstract}
Decision support systems for classification tasks are predominantly designed to predict the value of the ground truth labels.
However, since their predictions are not perfect, these systems also need to make human experts understand when and how to use these predictions to update their own predictions. 
Unfortunately, this has been proven challenging.
In this context, it has been recently argued that an alternative type of decision support systems may circumvent this challenge. 
Rather than providing a single label prediction, these systems provide a set of label prediction values constructed using a conformal predictor, namely a prediction set, and forcefully ask experts to predict a label value from the prediction set.
However, the design and evaluation of these systems have so far relied on stylized expert models, questioning their promise.
%
In this paper, we revisit the design of this type of systems from the perspective of
online learning and develop a methodology 
that does not require, nor assumes, an expert model.
%
Our methodology leverages the nested structure of the prediction sets provided by any conformal predictor and a natural counterfactual monotonicity assumption 
to achieve an exponential improvement in regret in comparison to vanilla bandit algorithms. 
We conduct a large-scale human subject study ($n = $ \nparticipants) to 
compare our methodology to several competitive baselines.
The results show that, for decision support systems based on prediction sets, limiting experts'{} level of agency 
leads to greater performance than 
allowing experts to always exercise their own agency.
We have made available the data gathered in our human subject study as well as an open source implementation of our system at \url{https://github.com/Networks-Learning/counterfactual-prediction-sets}.
\end{abstract}

\vspace{-2mm}
\section{Introduction}\label{sec:intro}
\vspace{-2mm}
Throughout the years, one of the main focus in the area of machine learning for decision support has been classification tasks. 
In this setting, the decision support system typically uses a classifier to predict the value of a ground truth label of interest and a human expert uses the predicted value to update their own prediction~\citep{bansal2019beyond, lubars2019ask, bordt2020humans}.
Classifiers have become notably
accurate in a variety of application domains such as medicine~\citep{jiao2020deep}, education~\citep{whitehill2017mooc}, or criminal justice~\citep{dressel2018accuracy}, to name a few.
However, their data-driven predictions are not always perfect~\citep{raghu2019algorithmic}.
%
As a result, there has been a flurry of work on helping human experts understand when and how to use the
predictions provided by these systems to update their own~\citep{papenmeier2019model, wang2021explanations, vodrahalli2022uncalibrated, liu2023learning}.
Unfortunately, it is yet unclear how to guarantee that, by using these systems, experts never decrease the average accuracy of their own predictions~\citep{yin2019understanding,zhang2020effect,suresh2020misplaced,lai2021towards}.

\begin{figure*}[t]
\centering
\includegraphics[width=.8\linewidth]{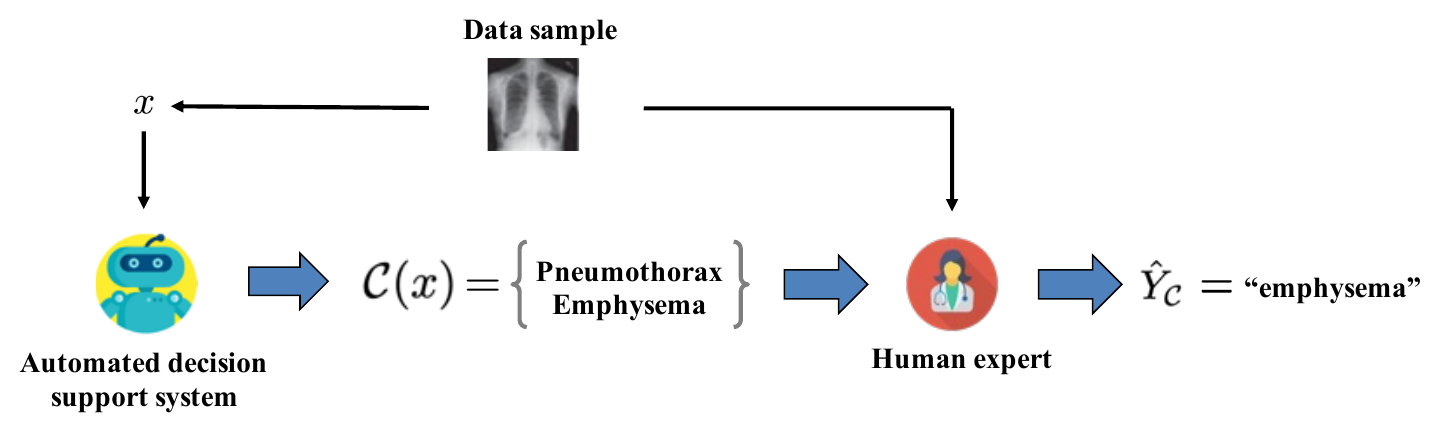}
\caption{Our automated decision support system $\Ccal$. 
Given a sample with a feature vector $x$, the system $\Ccal$ helps the expert by automatically 
narrowing down the set of potential label values to a subset of them $\Ccal(x) \subseteq \Ycal$, which
we refer to as a prediction set, using a set-valued predictor. 
The system forcefully asks the expert to predict a label value $\hat y_{\Ccal}$ from the prediction set $\Ccal(x)$, \ie, $\hat y_{\Ccal} \in \Ccal(x)$.
}
\label{fig:decision-support-system}
%
\end{figure*}

Very recently,~\citet{straitouri23improving} have argued that 
an alternative type of decision support systems may provide such a guarantee, by design.
Rather than providing a label prediction and letting human experts decide when and how to use the predicted label to update their own prediction, 
this type of systems provide a set of label predictions, namely a prediction set, and forcefully ask the experts to predict a label value from the prediction set, as shown in Figure~\ref{fig:decision-support-system}.\footnote{There are many systems used everyday by experts that, under normal operation, limit experts'{} level of agency. For example, think of a pilot who is flying a plane. There are automated, adaptive systems that prevent the pilot from taking certain actions based on the monitoring of the environment.}
Their key argument is that, if the prediction set is constructed using conformal prediction~\citep{VovkBook, angelopoulos2021gentle}, then one can precisely trade-off the probability that the ground truth label is not in the prediction set, which determines how frequently the systems will mislead human experts\footnote{Since these systems do not allow experts to predict a label value if it lies outside the prediction set, if the prediction set does not contain the ground truth label, we know that the expert'{}s prediction will be incorrect.}, and 
the size of the prediction set, which determines the difficulty of the classification task the experts need to 
solve using the system.
%
%
However, Straitouri et al. are only able to find the (near-)optimal conformal predictor that maximizes average accuracy under the assumption that the experts'{} predictions follow a stylized expert model, which they also
use for evaluation.
%
In this work, 
our goal is to lift this assumption and 
efficiently find the optimal conformal predictor that 
maximizes the average accuracy achieved by real experts
using such a system.
\subsection{Our Contributions}
%
We start by formally characterizing experts'{} predictions over prediction sets 
constructed using a conformal predictor using a structural causal model (SCM)~\citep{pearl2009causal}.
Building upon this characterization, we identify the following natural 
counterfactual monotonicity assumption on the experts'{} predictions, which may be of independent interest.
If an expert succeeds (fails) at predicting the ground truth label from 
a prediction set and the set contains the ground truth label, 
the expert would have also succeeded (failed) had the prediction set been 
smaller (larger) but had still contained the ground truth label.
%
Then, we use this counterfactual monotonicity assumption and the nested 
structure of the prediction sets provided by conformal prediction to design
very efficient bandit algorithms to find the optimal conformal predictor.
%
In particular, we formally show that, in our setting, a variant of the successive elimination 
algorithm~\citep{slivkins2019introduction}, which we refer to as counterfactual 
successive elimination, achieves an exponential improvement in regret in comparison 
with vanilla successive elimination. 

Finally, we conduct a large-scale user study with \nparticipants~human 
subjects who make \npredictions~predictions over $19{,}200$ different 
pairs of natural images and prediction sets.
In our study, we experiment both with a \emph{strict} and a \emph{lenient} 
implementation of our decision support systems.
Under the strict implementation, experts can only predict a label 
value from the prediction set whereas, 
under the lenient implementation, 
experts are encouraged to predict a label value from the prediction 
set but have the possibility to predict other label values.
Perhaps surprisingly, our results demonstrate that, under the strict 
implementation, 
experts achieve higher accuracy.
This suggests that, for decision support systems based on prediction sets, limiting experts'{} level of agency 
leads to greater performance than 
allowing experts to always exercise their own agency.
%
%
Further, our results also demonstrate that, for the strict implementation,
%
%
the conformal predictor found by our counterfactual successive elimination algorithm offers greater performance than that found by the algorithm by~\citet{straitouri23improving}.

An open-source implementation of both the strict and the lenient implementation of our system as well as all the data gathered in our human subject study, which we refer to as ImageNet16H-PS, are available at \url{https://github.com/Networks-Learning/counterfactual-prediction-sets}.

\subsection{Further Related Work}
Our work builds upon further related work on set-valued predictors, multi-armed bandits, and learning under algorithmic triage.

%
%
Conformal predictors are just one among many different types of set-valued predictors~\citep{chzhen2021set}, \ie, predictors that, for each sample, output a set of label values.
In our work, we opted for conformal predictors over alternatives such as, \eg, reliable or cautious classifiers~\citep{yang2017cautious,mortier2021efficient,ma2021partial,nguyen2021multilabel}, because of their provable coverage guarantees, which allow us to control how frequently our decision support systems will mislead human experts.
Except for two notable exceptions by~\citet{straitouri23improving} and~\citet{babbar2022utility}, set-valued predictors have not been specifically designed to serve decision support systems.
%
Within these two exceptions, the work by~\citet{straitouri23improving} is
more related to ours. 
However, 
it assumes
%
that the experts'{} predictions are sampled from a multinomial logit model (MNL), a classical discrete choice model~\citep{heiss2016discrete}.
On the contrary, in our work, we do not make any parametric
assumptions about the distribution the experts’ predictions are sampled from.
%
The work by~\citet{babbar2022utility} studies the lenient implementation of our decision support systems, under which the experts appear to achieve lower accuracy in our human subject study, as shown in Figure~\ref{fig:acc-vs-alpha}. 
%
However, it considers a conformal predictor with a given coverage probability, rather than optimizing across conformal predictors.\footnote{\citet{babbar2022utility} 
reduce the size of the prediction sets constructed using conformal prediction by 
deferring some samples to human experts during calibration and testing. 
However, since such an optimization can be applied both to the strict and the lenient implementation of our system, and we do not find any reason why it would change the conclusions of our human subject study, for simplicity, we decided not to apply it.}

%
Within the vast literature of multi-armed bandits (refer to~\citet{slivkins2019introduction} for a recent review),  
our work is most closely related to causal bandits~\citep{lattimore2016causal,lee2018structural,de2020causal} and combinatorial multi-armed bandits~\citep{chen2013combinatorial}. 
%
%
In causal bandits, there is a known causal relationship between arms and rewards, similarly as in our work. 
However, the focus is on using this causal relationship,
rather than counterfactual inference, to explore more efficiently and achieve lower regret.
%
In combinatorial multi-armed bandits, one can pull any subset of arms, namely a super arm, at the same time. Then, the goal is to identify the (near-)optimal super arm.
While one could view our problem as an instance of combinatorial multi-armed bandits, our goal is not to identify the optimal super arm but the optimal base arm.
Further, our work also relates to an extensive line of work using multi-armed bandits in real-world applications, which includes, but is not limited to, efficient data collection for clinical trials~\citep{durand2018contextual}, dynamic product pricing~\citep{misra2019dynamic, mueller2019low}, optimal assortment selection~\citep{agrawal2019mnl}, and anomaly detection on networks~\citep{ding2019interactive}. The survey by~\citet{bouneffouf2020survey} includes more examples of real-world applications where multi-armed bandits have been used.

In learning under algorithmic triage, a classifier predicts the ground truth label of a given fraction of the samples and leaves the remaining ones to a human expert, as instructed by a triage policy~\citep{mozannar2020consistent, de2020regression, de2021classification, okati2021differentiable}. 
In contrast, in our work, for each sample, a classifier is used to 
build a set of label predictions and the human expert needs to predict 
a label value from the set.

\vspace{-2mm}
\section{Decision Support Systems based on Prediction Sets}\label{sec:setting}
\vspace{-2mm}
Given a multiclass classification task where, for each sample, a human expert needs to predict
a label $y \in \Ycal = \{1, \ldots ,L \}$, with a feature vector
$x \in \Xcal$, with $x, y \sim P(X, Y)$,
the decision support system $\Ccal: \Xcal \rightarrow 2^{\Ycal}$ helps the expert by automatically
narrowing down the set of potential label values to a subset of them $\Ccal(x) \subseteq \Ycal$,
which we refer to as a prediction set, using a set-valued predictor~\citep{chzhen2021set}.
Here, for reasons that will become apparent later, we focus on a strict implementation of the system that, for any $x \in \Xcal$, forcefully asks the expert'{}s prediction $\hat y \in \Ycal$, with $\hat y \sim P(\hat{Y}_{\Ccal} \given X, \Ccal(X))$, to belong to the prediction set $\Ccal(x)$.
More formally, this is equivalent to assuming that $P(\hat{Y}_{\Ccal} = \hat y \given X = x, \Ccal(x)) = 0$ 
for all $\hat y \notin \Ccal(x)$.
Refer to Figure~\ref{fig:decision-support-system} for an illustration of the decision support system $\Ccal$. 

Then, the goal is to find the optimal decision support system $\Ccal^{*}$ that maximizes the 
average accuracy of the expert'{}s predictions, \ie, 
\begin{equation*}
\Ccal^{*} = \argmax_{\Ccal}\, \EE_{X ,Y ,\hat{Y}_{\Ccal}}[\II\{\hat{Y}_{\Ccal}=Y\}],
\end{equation*}
where $X,Y$$\sim$$P(X, Y)$ and $\hat{Y}_{\Ccal}$$\sim$$P(\hat{Y}_{\Ccal} \given X, \Ccal(X))$.
%
However, to solve the above maximization problem, we need to first specify the class of set-valued 
predictors we aim to maximize average accuracy upon.
Here, we favor conformal predictors~\citep{VovkBook, angelopoulos2021gentle} over alternatives for two key reasons.
First, they provide prediction sets with a nested structure that, together with the counterfactual monotonicity assumption, will allow us to find the optimal conformal predictor that maximizes the average accuracy of the expert'{}s predictions very efficiently. 
Second, they allow for a precise control of the trade-off between how frequently the expert is misled by the system and the difficulty of the classification task she needs to solve, as we discuss next.

Given a user-specified parameter $\alpha \in [0, 1]$, 
a conformal predictor uses a pre-trained classifier $\hat{f}(x) \in [0, 1]^{L}$
and a calibration set $\Dcal_{\text{cal}} = \{(x_i,y_i)\}_{i=1}^{m}$, where $(x_i, y_i) \sim P(X,Y)$, 
to construct the prediction sets $\Ccal_{\alpha}(X)$ as follows:\footnote{The assumption that $\hat{f}(x) \in [0,1]^{L}$ is without loss of generality. Here, the higher the score $\hat {f}_y(x)$, the more the classifier believes the ground truth label $Y = y$.}
\begin{equation} \label{eq:prediction-set}
\Ccal_{\alpha}(X) = \{y \given s(X, y) \leq \hat{q}_{\alpha} \},
\end{equation}
where $s(x_i, y_i) = 1 - \hat{f}_{y_i}(x_i)$ is called the conformal score\footnote{In general, the 
conformal score $s(x,y)$ can be any function of $x$ and $y$ measuring the \emph{similarity} between 
samples. Here, we choose $s(x,y) = 1 - \hat{f}_{y}(x)$ following \citet{sadinle2019least}.},
$\hat{f}_{y_i}(x_i)$ is the output of the classifier (\eg, the softmax score) for feature vector $x_i$ and label value $y_i$,
and 
$\hat{q}_{\alpha}$ is the $\frac{\lceil(m + 1)(1 - \alpha)\rceil}{m}$ empirical quantile of the conformal 
scores $s(x_1, y_1), \ldots, s(x_m, y_m)$.
Here, note that, for any sample with feature vector $x$, the prediction sets are nested with respect to the parameter $\alpha$, \ie,
$\Ccal_{\alpha}(x) \subseteq \Ccal_{\alpha'}(x)$ for any $\alpha > \alpha'$.
%
%
Moreover, 
the conformal predictor enjoys probably approximately correct (PAC) coverage guarantees, \ie, given tolerance values $\delta, \epsilon \in (0, 1)$, we can compute the minimum size $m$ of the calibration set $\Dcal_{\text{cal}}$ such that, with probability $1-\delta$, it holds that~\citep{vovk2012conditional} 
\begin{equation*}
1 - \alpha - \epsilon \leq \PP[Y \in \Ccal_{\alpha}(X) \given \Dcal_{\text{cal}}] \leq 1 - \alpha + \epsilon,
\end{equation*}
where $(1-\alpha)$ is called the (user-specified) coverage probability.\footnote{Most of the literature on conformal prediction focuses on marginal coverage rather than PAC coverage. 
However, since we will optimize the performance of our system with respect to $\alpha$, 
we cannot afford marginal coverage guarantees, as discussed in~\citet{straitouri23improving}.}
%
%
Then, we can conclude that, with high probability, for a fraction $(1-\alpha)$ of the samples, $\Ccal_{\alpha}(x)$ 
contains the ground truth label and thus cannot mislead the expert---if the expert would succeed at 
predicting the ground truth label $y$ of a sample with feature vector $x$ on her own, 
%
she could still succeed using $\Ccal_{\alpha}$ because $\Ccal_{\alpha}(x)$ contains the ground truth 
label. 
On the flip side, for a fraction $\alpha$ of the samples, we know that, if the expert uses $\Ccal_{\alpha}$, she will fail at predicting the ground truth label.\footnote{There may be some feature vectors $x$ for which, in principle, a conformal predictor may return an empty prediction set $\Ccal_{\alpha}(x)$. However, during deployment, one can trivially conclude that they will not contain the ground truth label. Hence, in those cases, $\Ccal_{\alpha}$ may allow the expert to choose from $\Ycal$ (or any other subset of labels).} 
Further, we know that the smaller (larger) the value of $\alpha$, the larger (smaller) the size of $\Ccal_{\alpha}(x)$, and thus the higher (lower) the difficulty of the classification task the expert must solve~\citep{wright1977phased,beach93, BENAKIVA19959}.

The important point above is that, since $\alpha$ is a parameter we choose, we can precisely control 
the trade-off between how frequently the system misleads the expert and the difficulty of the classification tasks the expert needs to solve.
Finally, note that, if we do not forcefully ask the expert to predict a label value from the prediction set, we would not be able to have this level of control and good performance would depend on the expert developing a good sense on when to predict a label from the prediction set and when to predict a label 
from outside the set. 

\vspace{-2mm}
\section{Prediction Sets through a Causal Lens}\label{sec:causal-model}
\vspace{-2mm}
In this section, we start by characterizing how human experts make predictions using a decision
support system via a structural causal model (SCM)~\citep{pearl2009causal}, which we denote as 
$\Mcal$. 
Our SCM $\Mcal$ is defined by the following assignments:
\begin{align} \label{eq:scm}
    \Ccal_{\mathrm{A}}(X) = f_{\Ccal}(X, \mathrm{A}), \quad \hat{Y}_{\Ccal_{\mathrm{A}}} = f_{\hat Y}(U, V, \Ccal_{\mathrm{A}}(X)), \quad  X = f_{X}(V) \quad \text{and} \quad Y = f_Y(V)
\end{align}
where $\mathrm{A}$, $U$ and $V$ are independent exogenous random variables and $f_{\Ccal}$, $f_{\hat Y}$, $f_{X}$ and $f_{Y}$ are given functions.
The exogenous variables $\mathrm{A}$, $U$ and $V$ characterize
the (user-specified) coverage 
probability,
the expert'{}s individual characteristics,
and the data generating process for the feature vectors $X$ and ground truth labels $Y$, respectively.
The function $f_{\Ccal}$ is a set-function directly defined by the conformal predictor, \ie, $f_{\Ccal}(X = x, A = \alpha) = \Ccal_{\alpha}(x)$,
where the calibration set $\Dcal_{\text{cal}}$ is given and thus it does not appear explicitly as an
independent variable.
Further, as argued elsewhere~\citep{pearl2009causal}, we can always find a distribution for the exogenous variables $U$ and $V$ and functions $f_{X}$, $f_{Y}$ and $f_{\hat Y}$ such that the observational distributions $P(X, Y)$ and $P(\hat Y_{\Ccal_{\alpha}} \given X, \Ccal_{\alpha}(X))$ of interest, defined in the previous section, are given by the distribution $P^{\Mcal}$ entailed by the SCM $\Mcal$, \ie, $P(X=x, Y = y) = P^{\Mcal}(X=x, Y = y)$ and
\begin{align*}
    P(\hat Y_{\Ccal_{\alpha}} = \hat{y} \given X = x, \Ccal_{\alpha}(X) = \Ccal_{\alpha}(x)) =  P^{\Mcal\,;\, \text{do}(\Ccal_{\mathrm{A}}(X) = \Ccal_{\alpha}(x))}(\hat Y_{\Ccal_{\mathrm{A}}} = \hat{y} \given X = x),
\end{align*}
where $\text{do}(\Ccal_{\mathrm{A}}(X) = \Ccal_{\alpha}(x))$ denotes a (hard) intervention in which
the first assignment in Eq.~\ref{eq:scm} is replaced by the value $\Ccal_{\alpha}(x)$. 
Here, note that we model the stochasticity in both features $X$ and labels $Y$ through the exogenous random variable $V$, instead of considering $X = f_X(V)$ and $Y = f_Y(X)$, to allow for both causal and anticausal features $X$~\citep{scholkopf2012causal}. 
In what follows, we will use the above SCM $\Mcal$ to formally reason about the predictions 
$\hat Y_{\Ccal_{\alpha}}$ made by a human expert under different support systems $\Ccal_{\alpha}$ 
and then introduce two natural monotonicity assumptions from first principles.

Given a sample with feature vector $x$, we can first conclude that, since the expert's predictions
only depend on the (user-specified) coverage probability $(1-\alpha)$ through the prediction set 
$\Ccal_{\alpha}(x)$, 
it must hold that, for any pair of decision support systems $\Ccal_{\alpha}$ and $\Ccal_{\alpha'}$ such that 
$\Ccal_{\alpha}(x) = \Ccal_{\alpha'}(x)$, if we observe that the expert has predicted $\hat{Y}_{\Ccal_{\alpha}} = \hat{y}$ using $\Ccal_{\alpha}$, 
we can be certain that she would have predicted $\hat{Y}_{\Ccal_{\alpha'}} = \hat y$ had she used 
$\Ccal_{\alpha'}$ while holding ``everything else fixed''~\citep{pearl2009causal}.
%
Next, motivated by prior empirical studies in the psychology and marketing literature~\citep{schwartz2004paradox, haynes2009testing, kuksov2010more, chernev2015choice}, which suggest 
that increasing the number of alternatives in a decision making task increases its difficulty, we 
hypothesize and later on empirically verify (refer to Figure~\ref{fig:acc-set-size-across-experts} in Appendix~\ref{app:monotonicity}) that the following interventional monotonicity assumption holds: 
\vspace{-1mm}
\begin{assumption}[Interventional monotonicity]\label{asm:interventional-monotonicity}
The experts' predictions satisfy interventional monotonicity if and only if, for any $x \in \Xcal$ and any 
$\Ccal_{\alpha}$ and $\Ccal_{\alpha'}$ such that $Y \in \Ccal_{\alpha}(x) \subseteq \Ccal_{\alpha'}(x)$, 
it holds that
\begin{align} \label{asm:success-probability-monotonicity}
    P^{\Mcal\,;\, \text{do}(\Ccal_{\mathrm{A}}(X) = \Ccal_{\alpha}(x))}(\hat Y_{\Ccal_{\mathrm{A}}} = Y \given X = x) \geq 
    P^{\Mcal\,;\, \text{do}(\Ccal_{\mathrm{A}}(X) = \Ccal_{\alpha'}(x))}(\hat Y_{\Ccal_{\mathrm{A}}} = Y \given X = x),
\end{align}
where the probability is over the uncertainty on the expert'{}s individual characteristics
and the data generating process.
\end{assumption}
%
%
Moreover, motivated by the aforementioned empirical stu\-dies, we further hypothesize that the following natural counterfactual monotonicity assumption, which is a suffi\-cient condition for interventional 
monotonicity, 
may also hold:
\begin{assumption}[Counterfactual monotonicity]\label{asm:reward-monotonicity}
The experts' predictions sa\-tis\-fy counterfactual mo\-no\-to\-ni\-ci\-ty if and only if, for any $x \in \Xcal$ and any $\Ccal_{\alpha}$ and $\Ccal_{\alpha'}$ such that $Y \in \Ccal_{\alpha}(x) \subseteq \Ccal_{\alpha'}(x)$, 
it holds that 
\begin{equation*}
\II\{ f_{\hat Y}(u, v, \Ccal_{\alpha}(x)) = Y \} \geq \II\{ f_{\hat Y}(u, v, \Ccal_{\alpha'}(x)) = Y \}
\end{equation*}
for any $u \sim P^{\Mcal}(U)$ and $v \sim P^{\Mcal}(V \given X = x)$.
\end{assumption}
\vspace{-1mm}
The above assumption is a sufficient condition for interventional monotonicity since we can obtain the interventional monotonicity inequality given by  Eq.~\ref{asm:success-probability-monotonicity}, by taking the expectation in both sides of the counterfactual monotonicity inequality, \ie, 
$\II\{ f_{\hat Y}(u, v, \Ccal_{\alpha}(x)) = Y \} \geq \II\{ f_{\hat Y}(u, v, \Ccal_{\alpha'}(x)) = Y \}$
 over $u, v$.
Further, the counterfactual monotonicity assumption directly implies that, for any sample with feature vector $x$ and any $\Ccal_{\alpha}$ and $\Ccal_{\alpha'}$ such that $Y \in \Ccal_{\alpha}(x) \subseteq \Ccal_{\alpha'}(x)$, 
if we observe that an expert has succeeded at predicting the ground truth label $Y$ using $\Ccal_{\alpha'}$, she would have also succeeded had she used $\Ccal_{\alpha}$ and, conversely, if she has failed at predicting $Y$ using $\Ccal_{\alpha}$, she would have also failed had she used $\Ccal_{\alpha'}$, while holding ``everything else fixed''.
In other words, under the counterfactual monotonicity assumption, the counterfactual dynamics of the expert under certain alternative prediction sets are identifiable and purely deterministic. 
However, note that this does not prevent the factual dynamics of the expert from being stochastic and fallible, as extensively argued in psychology and behavior literature~\citep{kahneman1973psychology, tversky1981framing, kahneman1984choices, loewenstein1992anomalies, loewenstein2003role}.

In what follows, we will leve\-rage this assumption to develop very efficient online algorithms to find the optimal conformal predictor among those using a given calibration set $\Dcal_{\text{cal}}$, \ie, $\alpha^{*} = \argmax_{\alpha} \EE_{X,Y,\hat{Y}_{\Ccal_\alpha}}[\II\{\hat{Y}_{\Ccal_{\alpha}} = Y\}]$.

\xhdr{Remarks}
Since the counterfactual monotonicity assump\-tion lies within level three in the ``ladder of causa\-tion''~\citep{pearl2009causal}, we cannot validate it using observational nor interventional experiments.
However, the good practical performance of our online algorithms in our human subject study, shown in Figures~\ref{fig:regret} and~\ref{fig:acc-vs-alpha}, suggests that it may hold 
in the context of the prediction task we have considered. 
Moreover, in Appendix~\ref{app:sensitivity-montonicity}, we carry out an additional 
sensitivity analysis,
which shows that the performance of our algorithms degrades gracefully with 
respect to the amount of violations of the counterfactual monotonicity assumption.

\vspace{-2mm}
\section{Finding the Optimal Conformal Predictor using Counterfactual Prediction Sets}\label{sec:successive-elimination}
\vspace{-2mm}
Given a fixed calibration set $\Dcal_{\text{cal}} = \{(x_i, y_i)\}_{i=1}^{m}$, there exist only $m$ different conformal predictors. 
This is because the empirical quantile $\hat{q}_{\alpha}$, which the subsets $\Ccal_{\alpha}(x_i)$
depend on, can only take $m$ different values~\citep{straitouri23improving}.
As a result, to find the optimal conformal predictor, we just need to solve the following 
maximization problem:
\vspace{-2mm}
\begin{equation} \label{eq:optimal-alpha}
     \alpha^{*} = \argmax_{\alpha \in \Acal} \, \EE_{X,Y,\hat{Y}_{\Ccal_{\alpha}}} \left[ \II\{\hat Y_{\Ccal_{\alpha}} = Y\} \right],
\end{equation}
where $\Acal = \{\alpha_i\}_{i \in [m]}$, with $\alpha_i = 1 - i/(m+1)$ and $[m] = \{1, 2, \dots, m\}$.
However, since we do not know the causal mechanism experts use to make predictions over prediction sets, 
we need to trade-off exploitation, 
\ie, maximizing the expected accuracy, 
and exploration, 
\ie, learning about the accuracy achieved by the experts under each conformal predictor. To this end, we 
look at the problem from the perspective of multi-armed bandits~\citep{slivkins2019introduction}.

In our problem, each arm corresponds to a different parameter value $\alpha$ and, 
at each round $t$, 
a (potentially different) human expert receives a sample with feature vector $x_t$, 
picks a label value $\hat{y}_t$ from the prediction set $\Ccal_{\alpha_t}(x_t)$ provided by the 
conformal predictor with $\alpha_t \in \Acal$,
and obtains a reward $\II\{ \hat{y}_t = y_t \} \in \{0, 1\}$.
Here, we observe $x_t$ at the beginning of each round and $y_t$ and $\II\{ \hat{y}_t = y_t \}$ at the end of each round.
Then, the goal is to find a sequence of parameter values $\{ \alpha_t \}_{t=1}^{T}$
with desirable properties
in terms of total regret $R(T)$, which is given by:
\vspace{-2mm}
\begin{align}\label{eq:regret}
    R(T) = T\cdot\EE_{X, Y, \hat{Y}_{\Ccal_{\alpha^{*}}}}\left[\II\{ \hat{Y}_{\Ccal_{\alpha^{*}}} = Y \}\right] -  \sum_{t=1}^{T} \EE_{X, Y, \hat{Y}_{\Ccal_{\alpha_t}}}\left[\II\{ \hat{Y}_{\Ccal_{\alpha_t}} = Y \}\right],
\end{align}
where $\alpha^{*}$ is the optimal parameter value, as defined in Eq.~\ref{eq:optimal-alpha}.
At this point, one could think of resorting to any of the well-known algorithms from the literature 
on stochastic multi-armed bandits~\citep{slivkins2019introduction}, such as \texttt{UCB1} or successive 
elimination, to decide which arm to pull, \ie, which $\alpha_t$ to use, at each round $t$.
These algorithms would achieve an expected regret $\EE[R(t)] \leq O( \sqrt{m t \log T})$ 
for any $t \leq T$, where the expectation is over the randomness in the execution of the algorithms.
However, in our problem setting, we can do much better than that---in what follows, we will design an 
algorithm based on successive elimination that achieves an expected regret $\EE[R(t)] \leq O( \sqrt{t \log m  \log T})$ for any $t \leq T$.

The successive elimination algorithm keeps a set $\Acal_{\text{active}}$ of \emph{active} arms $\alpha$, 
which initially sets to $\Acal_{\text{active}} = \Acal$. 
Then, it pulls a different arm $\alpha \in \Acal_{\text{active}}$, without repetition, until it has pulled all arms in $\Acal_{\text{active}}$.
Assume it has pulled all arms at round $t$. Then, it computes an upper and a lower confidence bound on the average reward associated to each arm $\alpha$, 
$\texttt{UCB}_t(\alpha) = \hat{\mu}_{t}(\alpha) + \epsilon_t(\alpha)$ and $\texttt{LCB}_t(\alpha) = \hat{\mu}_{t}(\alpha) - \epsilon_t(\alpha)$, where
\begin{align*}
    \hat{\mu}_{t}(\alpha) = \frac{\sum_{t' \leq t} \II\{ \hat{y}_{t'} = y_{t'} \wedge \alpha_{t'} = \alpha \}}{\sum_{t' \leq t} \II\{ \alpha_{t'} = \alpha \}} \quad \text{and} \quad \epsilon_t(\alpha) = \sqrt{\frac{2 \log T}{\sum_{t' \leq t} \II\{ \alpha_{t'} = \alpha \}}},
\end{align*}
and \emph{deactivates} any arm $\alpha \in \Acal_{\text{active}}$ for which there exists $\alpha' \in \Acal_{\text{active}}$ such that $\texttt{UCB}_t(\alpha) < \texttt{LCB}_t(\alpha'{})$.
Then, it repeats the same procedure until the maximum number of rounds $T$ is reached or until $|\Acal_{\text{active}}| = 1$. 

The rate at which successive elimination deactivates arms and, in turn, the expected regret, is 
essentially limited by the fact that, to update $\hat{\mu}_t(\alpha)$ and $\epsilon_t(\alpha)$ for 
every arm $\alpha \in \Acal_{\text{active}}$, it needs to pull $O(m)$ arms.
However, in our problem setting, there exists an efficient strategy to update $\hat{\mu}_t(\alpha)$ and 
$\epsilon_t(\alpha)$ by pulling just $O(\log m)$ arms.
\begin{algorithm}[t]
\caption{Counterfactual Successive Elimination}
\begin{algorithmic}[t] \label{alg:counterfactual-successive-elimination}
\small
\STATE{\textbf{Input:} $\Acal, T, \Dcal_{\text{opt}}$}
\STATE{\textbf{Output:} $\Acal_{\text{active}}$}
\vspace{1mm}
\STATE{$\Acal_{\text{active}} \leftarrow \Acal$}
\STATE{$t \leftarrow 0$, $\gamma \leftarrow 0$, $\nu \leftarrow 0$}
\WHILE{$t<T \vee |\Acal_{\text{active}}| > 1$ }
    \STATE{$\Acal_{\text{unexplored}} \leftarrow \Acal_{\text{active}}$}
    \WHILE{$ \Acal_{\text{unexplored}} \neq \varnothing \wedge t < T$}
        \STATE{$\tilde{\alpha} \leftarrow \textsc{median}(\Acal_{\text{unexplored}} )$}
        \STATE{$(x_t, y_t)\sim\Dcal_{\text{opt}}$}
        \STATE{Deploy $\Ccal_{\tilde{\alpha}}(x_t)$ and observe $\II\{\hat{y}_t = y_t\}$ and $y_t$}
        \STATE{ $\Acal_{\text{unexplored}}, \gamma, \nu \leftarrow$  \textsc{Alg.\ref{alg:update-rewards}}$(\Acal_{\text{unexplored}}, \gamma, \nu, \tilde{\alpha}, x_t, y_t, \hat{y}_t)$ 
        }
        \STATE{$t \leftarrow t + 1$}
    \ENDWHILE
    \FOR{$\alpha \in \Acal_{\text{active}}$}
        \STATE $\mu(\alpha) = \gamma(\alpha)/\nu(\alpha)$
        \STATE $\epsilon(\alpha) = \sqrt{2 \log T / \nu(\alpha)}$
    \ENDFOR
    \FOR{$\alpha \in \Acal_{\text{active}}$}
        \IF[Apply deactivation rule]{$\exists \alpha' \in \Acal_{\text{active}} : \mu(\alpha) + \epsilon(\alpha) < \mu(\alpha') - \epsilon(\alpha')$}
            \STATE{$\Acal_{\text{active}} \leftarrow \Acal_{\text{active}} \setminus \{\alpha\}$}
        \ENDIF
    \ENDFOR
\ENDWHILE
\end{algorithmic}
\end{algorithm}

\begin{algorithm}[t]
\caption{It updates $\Acal_{\text{unexplored}}$, $\gamma$ and $\nu$}
\begin{algorithmic}[t] \label{alg:update-rewards}
\small
\STATE{\textbf{Input:} $\Acal_{\text{unexplored}}, \gamma, \nu, \tilde{\alpha}, x, y, \hat{y}$}
\STATE{\textbf{Output:} $\Acal_{\text{unexplored}}, \gamma, \nu$}

\vspace{1mm}
\STATE{$\alpha^{\dagger} \leftarrow \inf\{ \alpha : y \notin \Ccal_{\alpha}(x) \}$}
\FOR{$\alpha' \in \Acal_{\text{unexplored}} : \alpha' \geq \alpha^{\dagger}$}
    \STATE $\nu(\alpha') \leftarrow \nu(\alpha')+1$
\ENDFOR
\vspace{1mm}            
\IF{$\II\{\hat{y} = y\} = 0$}
            \IF{$y \in \Ccal_{\tilde{\alpha}}(x)$}            
                \FOR{$\alpha' \in \Acal_{\text{unexplored}} : \alpha' \leq \tilde{\alpha}$}
                    \STATE $\nu(\alpha') \leftarrow \nu(\alpha')+1$
                \ENDFOR
                
                \STATE{$\Acal_{\text{unexplored}} \leftarrow \Acal_{\text{unexplored}} \setminus \{\alpha': \alpha' \leq \tilde{\alpha}\}$}
            \ENDIF
        \ELSE
            \FOR{$\alpha' \in \Acal_{\text{unexplored}} : \tilde{\alpha} \leq \alpha' < \alpha^{\dagger}$}
                    \STATE $\nu(\alpha') \leftarrow \nu(\alpha')+1$
                    \STATE $\gamma(\alpha') \leftarrow \gamma(\alpha')+1$
            \ENDFOR
            \STATE{$\Acal_{\text{unexplored}} \leftarrow \Acal_{\text{unexplored}} \setminus \{\alpha' :  \alpha' \geq \tilde{\alpha}\}$}
        \ENDIF
        \STATE{\textbf{return} $\Acal_{\text{unexplored}}, \gamma, \nu$}
\end{algorithmic}
\end{algorithm}
In the first round, our algorithm pulls the arm whose corresponding parameter value $\tilde{\alpha}$ is the \emph{median}\footnote{The $\frac{m}{2}$-th largest value if $m$ is even or the $\frac{m+1}{2}$-th largest value if $m$ is odd.} of all values in $\Acal_{\text{active}}$. 
We distinguish two cases. 
First, assume that the expert has failed at predicting the ground truth label $y_1$ using $\Ccal_{\tilde{\alpha}}$, \ie, $\II\{ \hat{y}_1 = y_1 \} = 0$.
If $y_1 \notin C_{\tilde{\alpha}}(x_1)$, we know that she would have also 
failed had she used any $\Ccal_{\alpha'}$ such that $\alpha' > \tilde{\alpha}$ since $\Ccal_{\alpha'}(x) 
\subseteq \Ccal_{\tilde{\alpha}}(x)$. 
If $y_1 \in C_{\tilde{\alpha}}(x_1)$, we know that she would have also failed had she used any $\Ccal_{\alpha'}$ such that $\alpha' < \tilde{\alpha}$ due to the counterfactual monotonicity 
assumption.
Second, assume that the expert has succeeded at predicting $y_1$ using $\Ccal_{\tilde{\alpha}}$, 
\ie, $\II\{ \hat{y}_1 = y_1 \} = 1$.
Then, for any $\alpha' > \tilde{\alpha}$, we know that, if $y_1 \in \Ccal_{\alpha'}(x_1)$, the same expert would have also succeeded had she used $\Ccal_{\alpha'}$ due to the counterfactual monotonicity assumption and, if $y_1 \notin \Ccal_{\alpha'}(x_1)$, the expert would have trivially failed had she used $\Ccal_{\alpha'}$.
In both cases, the algorithm observes the reward for one arm and counterfactually infers the reward for at least one \emph{half}\footnote{One \emph{half} is $\frac{m}{2}$ or $\frac{m}{2} - 1$ values if $m$ is even and $\frac{m-1}{2}$ values if $m$ is odd.} of the arms in $\Acal_{\text{active}}$. 

In the next $O(\log m)$ rounds, it repeats the same reasoning, it pulls the arm whose parameter value is the median of the remaining (at most) \emph{half} of the arms whose reward has not yet observed or counterfactually inferred, until it has observed or counterfactually inferred the reward of all arms at least once.
Then, it computes 
\begin{equation*}
    \hat{\mu}_{t}(\alpha) = \frac{\sum_{t' \leq t} \gamma_{t'}(\alpha)}
    {\sum_{t' \leq t} \nu_{t'}(\alpha)} \quad \text{and} \quad
    \epsilon_t(\alpha) = \sqrt{\frac{2 \log T}{\sum_{t' \leq t} \nu_{t'}(\alpha)}}, 
\end{equation*}
where $\gamma_{t'}(\alpha) = \II\{ \hat{y}_{t'} = y_{t'} \wedge \alpha_{t'} \leq \alpha \wedge y_{t'} \in \Ccal_{\alpha}(x_{t'}) \}$ and
\begin{align*}
    \nu_{t'}(\alpha) = \gamma_{t'}(\alpha) 
    + \II\{ \hat{y}_{t'} \neq y_{t'} \wedge \alpha_{t'} \geq \alpha \wedge y_{t'} \in \Ccal_{\alpha_{t'}}(x_{t'}) \} + 
    \II\{ y_{t'} \notin \Ccal_{\alpha}(x_{t'}) \}
\end{align*}
and, similarly as standard successive elimination, it deactivates any arm $\alpha \in \Acal_{\text{active}}$ 
for which there exists $\alpha' \in \Acal_{\text{active}}$ such that $\texttt{UCB}_t(\alpha) < \texttt{LCB}_t(\alpha'{})$ and repeats the entire procedure until $T$ is reached or until $|\Acal_{\text{active}}| = 1$.
Algorithm~\ref{alg:counterfactual-successive-elimination} 
summarizes the overall algorithm, which we refer to
as counterfactual successive elimination (Counterfactual SE), and Theorem~\ref{thm:regret} below formalizes its regret guarantees (proven in Appendix~\ref{app:regret}):

\begin{theorem} \label{thm:regret}
Given a calibration set $\Dcal_{\text{cal}} = \{ (x_i, y_i) \}_{i=1}^{m}$ and a maximum number of rounds $T \geq \sqrt{m}$,
Counterfactual SE is guaranteed to achieve expected regret $\EE[R(t)] \leq O \left( \sqrt{t \log m \log T} \right)$ for any $t \leq T$.
\end{theorem}

Interestingly, one can use counterfactual rewards to improve other well-known bandit algorithms, not only successive elimination. 
For example, to use counterfactual rewards in \texttt{UCB1}, at each time step $t$, one pulls the arm $\alpha_{t} = \argmax_{\alpha \in \Acal}\texttt{UCB}_t(\alpha)$ and counterfactually infers the rewards for any $\alpha > \alpha_{t}$ or $\alpha < \alpha_{t}$, similarly as Counterfactual SE does for $\tilde{\alpha}$. 
In section~\ref{sec:experiments}, we evaluate the benefits of using counterfactual rewards both in successive elimination and in \texttt{UCB1}. 
Remarkably, our experimental results demonstrate that counterfactual \texttt{UCB1}, \ie, \texttt{UCB1} using counterfactual rewards, achieves lower expected regret than Counterfactual SE.
Motivated by this empirical finding, it would be very interesting, but challenging, 
to derive formal regret guarantees for counterfactual \texttt{UCB1} in future work. 
One of the main technical obstacles one would need to solve is that, in counterfactual \texttt{UCB1}, the number of counterfactually 
inferred rewards at each time step is unknown, whereas in Counterfactual SE, which at each time step counterfactually infers at least one half of the unobserved arms in $\Acal_{\text{active}}$. 

\xhdr{Remarks} Counterfactual SE assumes that, every time an expert predicts a sample, the ground truth label $y$ is (eventually) observed. 
However, there may be scenarios in which the ground truth labels are only observed during the design phase of the decision support system, but not during deployment.
In those cases, it may be more meaningful to look at the problem from the perspective of best-arm identification and derive theoretical guarantees regarding the probability that Counterfactual SE chooses the best arm~\citep{slivkins2019introduction}, which is left as future work.
In this context, it is worth noting that SE and \texttt{UCB1} offer desirable theo\-re\-ti\-cal guarantees for best-arm identification~\citep{even2006action,audibert2010best}.

\vspace{-2mm}
\section{Evaluation via Human Subject Study}\label{sec:experiments}
\vspace{-2mm}
In this section, we conduct a large-scale human subject study and show that:
\squishlist
\item[a)] human experts are more likely to succeed at predicting the ground truth label under smaller prediction sets, providing strong evidence that the interventional monotonicity assumption (Assumption~\ref{asm:interventional-monotonicity}) holds;
\item[b)] counterfactual successive elimination and counterfactual \texttt{UCB1} achieve a significant improvement in expected regret in comparison with their vanilla implementations; 
\item[c)] a strict implementation of our decision support system, which adaptively limits experts'{} level of agency, offers greater performance than a lenient implementation, which allows experts to always exercise their own agency.\footnote{All experiments ran on a Mac OS machine with an M1 processor and $16$GB memory.}
\squishend

\xhdr{Human subject study setup} To construct our dataset ImageNet16H-PS, we gathered \npredictions~label predictions from \nparticipants~human participants for $1{,}200$ unique images from the ImageNet16H dataset~\citep{steyvers2022bayesian} using Prolific.
Our experimental protocol received approval from the Institutional Review Board (IRB) at the University of Saarland,
each participant was rewarded with $9$\textsterling~per hour pro-rated, following Prolific'{}s payment principles, and consented to participate by filling a consent form that included a detailed description of the study processes,
and the collected data did not include any personally identifiable information.
Each image in the ImageNet16H dataset belongs to one of $16$ different categories (\eg, animals, vehicles as well as every day objects), which serve as labels.
%
In our study, we used always the same classifier, namely the pre-trained VGG-19~\citep{simonyan2014very} after $10$ epochs of fine-tuning as provided by~\citet{steyvers2022bayesian}  and a fixed calibration set of $120$ images, picked at random.\footnote{The model and the dataset ImageNet16H are publicly available at \url{https://osf.io/2ntrf/}.} The average accuracy of the classifier (over the images not in the calibration set) is $0.848$. 
For each image, we first computed all possible prediction sets that any conformal predictor using the above classifier and calibration set could construct.
Then, we created $715$ questionnaires, each with a set of images and, for each image, a multiple choice question using a prediction set.
Under the strict implementation of our system, the multiple choice options included only the label values of the corresponding prediction set and, under the lenient implementation, they additionally included an option ``Other'', which allowed participants to pick a label value outside the prediction set. 
Under both implementations, the questionnaires covered all possible pairs of images and prediction sets.
We provide additional information about the ImageNet16H dataset, further implementation details as well as screenshots of the questionnaires and the consent form in Appendix~\ref{app:experiment-details}.
In what follows, if not said otherwise, the results refer to the strict implementation of our system.

\begin{figure}[t]
    \centering
    \includegraphics[width=.6\linewidth]{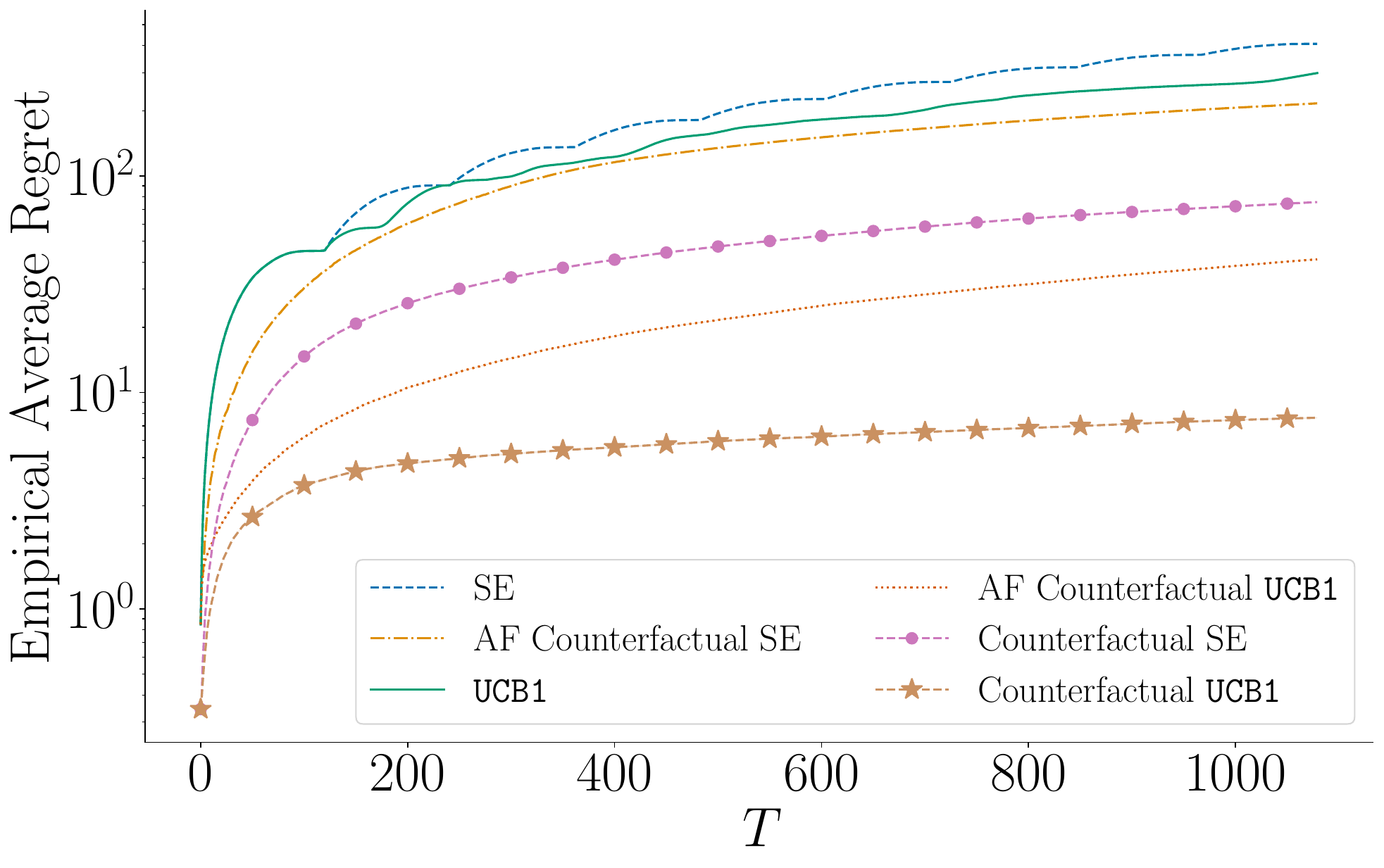}
    \caption{Empirical average regret achieved by six different bandit algorithms across $30$ different realizations. The standard error is not visible as it is always below $0.2$.
    }
    \label{fig:regret}
\end{figure}

\xhdr{Expert success probability vs.~prediction set size} As discussed previously, we cannot directly verify the counterfactual mo\-no\-to\-ni\-city assumption using an interventional study because it is a counterfactual property. However, we can verify interventional monotonicity, a necessary condition for counterfactual monotonicity to hold---whether, on average, experts 
are more likely to succeed at predicting the ground truth label using smaller prediction sets.
To this end, we stratify the images in the dataset with respect to their difficulty into groups and, for each group, we estimate the success probability per prediction set size averaged across all experts and across experts with the same level of competence. 
We consider groups of images, rather than single images, because we have too few expert predictions to derive reliable estimates of the success probability per image.
Refer to Appendix~\ref{app:experiment-details} for more details on how we stratify images and experts.
Figure~\ref{fig:acc-set-size-across-experts} in Appendix~\ref{app:monotonicity} shows that, as long as the images are not too easy, the experts are more accurate under smaller prediction sets---the interventional monotonicity assumption holds.

\xhdr{Regret analysis} In addition to validating the formal regret guarantees (Theorem~\ref{thm:regret}) of counterfactual successive elimination (Counterfactual SE), here,
we aim to evaluate the competitive advantage that counterfactual rewards bring to several bandit algorithms in terms of expected regret.\footnote{
The computation of the expected regret requires 
knowledge of $\alpha^{*}$. 
However, since we gathered expert predictions for each possible prediction set that any conformal predictor using the calibration set may construct, we could estimate the empirical success probability under each of them and find $\alpha^{*}$ by enumeration.
} 
To this end, we estimate the expected regret over a horizon of $1{,}080$ time steps over $30$ different realizations of the following algorithms:\footnote{Given that we have expert predictions using all possible prediction sets for $1{,}080$ images, not including the $120$ images in the calibration set, we can run any bandit algorithm \emph{faithfully} for $1{,}080$ time steps.}
a) vanilla successive elimination (SE), 
b) vanilla \texttt{UCB1}, 
c) counterfactual successive elimination (Counterfactual SE), 
d) counterfactual \texttt{UCB1}, 
e) assumption-free counterfactual successive elimination (AF Counterfactual SE), 
and f) assumption-free counterfactual \texttt{UCB1} (AF Counterfactual \texttt{UCB1}). 
The last two algorithms do not use the counterfactual monotonicity assumption but, for 
any sample $(x, y)$ and $\alpha \in \Acal$, they counterfactually infer that, 
for any $\alpha' \in \Acal$ such that $\alpha' \neq \alpha$ and $y \notin \Ccal_{\alpha'}(x)$, 
the expert would have failed to predict the ground truth label had she used $\Ccal_{\alpha'}$
and, for any $\alpha' \in \Acal$ such that $\Ccal_{\alpha'}(x) = \Ccal_{\alpha}(x)$, the expert
would have predicted the same label had she used $\Ccal_{\alpha'}$.
Figure~\ref{fig:regret} summarizes the results, which show that
counterfactual rewards provide a clear competitive advantage with respect to their vanilla counterparts and, by using the counterfactual monotonicity assumption, Counterfactual SE and Counterfactual \texttt{UCB1} are clear winners, suggesting that the assumption may (approximately) hold. 

\begin{figure}[t]
    \centering
    \includegraphics[width=0.6\linewidth]{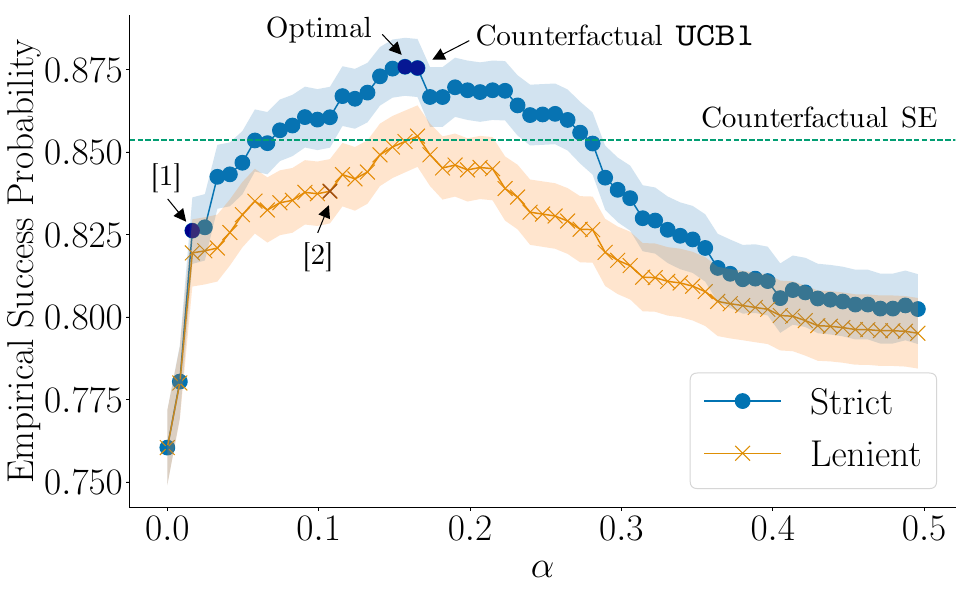}
    \caption{Empirical success probability achieved by all experts across all images using the strict and lenient implementation of our system $\Ccal_{\alpha}$ with different $\alpha$ values. 
    %
    For the strict implementation, we annotate the optimal $\alpha$ value, the $\alpha$ values found by the algorithms by~\citet{straitouri23improving} [1] and by counterfactual \texttt{UCB1}, as well as the average success probability achieved by the set of $\alpha$ values that remain active after running counterfactual SE. 
    For the lenient implementation, we annotate the $\alpha$ value used by~\citet{babbar2022utility} [2].
    The average accuracy of the classifier used by both the strict and the lenient implementation of our system is $0.848$ and the empirical success probability achieved by the experts on their own is $0.760$.
    The shaded areas correspond to a $95\%$ confidence interval.}
    \label{fig:acc-vs-alpha}
\end{figure}

\xhdr{Strict vs.~lenient implementation of our system} One of the key motivations to forcefully ask experts to predict label values from the
prediction sets provided by conformal prediction is to be able to trade-off how frequently
the system misleads experts and the difficulty of the task the expert needs to solve~\citep{straitouri23improving}. 
However, one could argue that a more lenient system, 
which allows experts to predict label values from outside the prediction sets as suggested by~\citet{babbar2022utility}, 
may offer greater performance since, in principle, it does not forcefully mislead 
an expert whenever the prediction set does not contain the ground truth label.
Here, we provide empirical evidence that suggests that this is not the case.
Figure~\ref{fig:acc-vs-alpha} demonstrates that a strict implementation of our 
system consistently offers greater performance than a lenient implementation 
across the full spectrum of competitive $\alpha$ values,\footnote{We have 
excluded values of $\alpha > 0.5$ to improve visibility, however, we include 
the full figure in Appendix~\ref{app:acc-vs-alpha}.} where we allow 
experts to predict any label value from $\Ycal$ whenever a prediction set is empty 
under both the strict and the lenient implementation of our system. In Appendix~\ref{app:acc-vs-alpha}, we further investigate why the strict implementation 
is superior.
Figure~\ref{fig:acc-vs-alpha} also demonstrates that counterfactual \texttt{UCB1} and counterfactual SE offer a significant advantage over the algorithm proposed by~\citet{straitouri23improving}, which uses a stylized expert model.

\vspace{-2mm}
\section{Discussion and Limitations}
\label{sec:discussion}
\vspace{-2mm}
In this section, we discuss several assumptions and limitations of our work, pointing out avenues for future research. 

\xhdr{Data}
We have assumed that the data samples and the expert predictions are drawn i.i.d. from a fixed distribution 
and the calibration set contains samples with noiseless ground truth labels. 
In future work, it would be very interesting to lift these assumptions.     
Regarding distribution shift, a good starting point may be the rapidly expanding literature on conformal prediction under distribution shift~\citep{tibshirani2020conformal,podkopaev2021distribution,gibbs2021adaptive};
regarding label noise, the recent work by~\citet{einbinder2022conformal}, which has found that conformal prediction is robust to label noise; and, regarding label ambiguity, 
the very recent work by~\citet{stutz2023conformal}.

Further, while the good empirical performance of both counterfactual SE and counterfactual \texttt{UCB1} in our human subject study suggest that the counterfactual monotonicity assumption may hold in the context of the classification task we have considered in the study, it would be important to investigate to what extent it holds in other classification tasks.
That said, we find it reassuring that the performance of counterfactual SE and counterfactual \texttt{UCB1} degrade gracefully with respect to the amount of violations of the counterfactual monotonicity assumption, as shown in Appendix~\ref{app:sensitivity-montonicity}.

\xhdr{Methodology}
We have adapted two well-known bandit algorithms---successive elimination and \texttt{UCB1}---so that they benefit from counterfactual rewards and, for successive elimination, we have theoretically shown that counterfactual rewards offer an exponential improvement in terms of regret.
It would be very interesting to adapt other well-known bandit algorithms, including Bayesian bandits algorithms
such as Thomson'{}s sampling, so that they also benefit from counterfactual rewards.
Moreover, we have focused on maximizing the average accuracy of the expert'{}s predictions. 
However, whenever the expert'{}s predictions are consequential to individuals, it would be important to extend our methodology to account for fairness considerations.

\xhdr{Decision making task} We have focused on designing decision support systems for multiclass classification tasks.
It would be very interesting to extend our approach and our notion of counterfactual monotonocity to other classification tasks.
To this end, a good starting point may be the framework of risk controlling prediction sets (RCPS) by~\citet{bates2021distributionfree}, which generalizes conformal prediction and does allow for a variety of classification tasks, including multilabel classification. 
Further, it would also be very interesting to investigate to what extend our ideas are useful in other types of decision tasks (\eg, reinforcement learning) and decision support systems (\eg, LLMs).

\xhdr{Human subject study} Our large-scale human subject study provides encouraging results and suggests our decision support system may be practical. 
However, it comprises only one classification task on a single benchmark dataset of natural images and one may question its generalizability.  
It would be important to conduct additional human subject studies in other real-world domains with domain experts (\eg, medical doctors).
However, it is worth highlighting that conducting such studies at scale would entail significant financial costs---the total cost of our human subject study, which did not rely on domain experts, was $7{,}150\pounds$.

\vspace{-2mm}
\section{Conclusions}
\vspace{-2mm}
We have looked at the development of decision support systems based on prediction sets for multiclass classification tasks from the perspective of online learning and counterfactual inference.
This perspective has allowed us to design a methodology that does not require, nor assumes, a stylized human expert model, and has modest computational and data requirements.
In doing so, we have also identified two natural monotoni\-ci\-ty assumptions, \ie, interventional monotonicity and counterfactual monotonicity, which may be of independent interest.
Further, we have conducted a large-scale human subject study that shows that our methodology is superior to the state of the art and, for decision support systems based on prediction sets, adaptively limiting experts' level of agency leads to greater performance.

\section*{Acknowledgements}
Gomez-Rodriguez acknowledges support from the European Research Council (ERC) under the European Union'{}s Horizon 2020 research and innovation programme (grant agreement No. 945719).

\section*{Impact Statement}
Our work has focused on maximizing the average accuracy achieved by a human expert using a decision support system based on conformal prediction.
However, whenever the expert'{}s predictions are consequential to individuals, it would be important to extend our methodology to account for fairness considerations.
In Section~\ref{sec:discussion}, we have also discussed other assumptions and limitations of our work, which may influence its impact in practice.

\bibliography{counterfactual_prediction_sets}
\bibliographystyle{apalike}

\newpage
\onecolumn

\appendix
\section{Proof of Theorem~\ref{thm:regret}} \label{app:regret}
Let us define the clean event $\Ecal = \{ | \hat{\mu}_{t}(\alpha) - \EE[ \II\{\hat{Y}_{\Ccal_{\alpha}} = Y\} ] |  \leq \epsilon_t(\alpha)|, \forall \alpha \in \Acal, \forall t \leq T \}$, with $\epsilon_t(\alpha) = \sqrt{2\log(T) / \nu_t(\alpha)}$, and decompose the average regret with respect to $\Ecal$ as follows:
 \begin{equation}\label{eq:exp-regret-clean-event}
     \EE[R(t)] = \EE[R(t) \given \Ecal] \PP[\Ecal] + \EE[R(t) \given \Bar{\Ecal}]  \PP[\Bar{\Ecal}], 
 \end{equation}
 where $\Bar{\Ecal}$ is the complement of $\Ecal$. 
Then, following~\citet{slivkins2019introduction}, we first assume $\Ecal = \{ | \hat{\mu}_{t}(\alpha) - \EE[ \II\{\hat{Y}_{\Ccal_{\alpha}} = Y\} ] |  \leq \epsilon_t(\alpha)|, \forall \alpha \in \Acal, \forall t \leq T \}$ holds and then show that the probability that $\Bar{\Ecal}$ holds is negligible.

Let $\alpha$ be any suboptimal arm and $\alpha^*$ be the optimal one, \ie, $\EE[ \II\{\hat{Y}_{\Ccal_{\alpha}} = Y\} ] < \EE[ \II\{\hat{Y}_{\Ccal_{\alpha^{*}}} = Y\} ]$. 
Let $t' \leq T$ be the last round in which we have applied the deactivation rule and $\alpha$ 
was active. 
Until then, both $\alpha$ and $\alpha^{*}$ are active and, as a result, in each phase we
collected a reward for each of them. 
Therefore, it must hold that $\nu_{t'}(\alpha) = \nu_{t'}(\alpha^{*})$ and thus $\epsilon_{t'}(\alpha^{*}) = \epsilon_{t'}(\alpha)$. 
Further, since, by assumption, $\Ecal$ holds, we have that: 
\begin{equation}\label{eq:delta-alpha}
    \Delta(\alpha) = \EE[ \II\{\hat{Y}_{\Ccal_{\alpha^{*}}} = Y\} ] -  \EE[ \II\{\hat{Y}_{\Ccal_{\alpha}} = Y\} ] \leq 2(\epsilon_{t'}(\alpha^{*}) + \epsilon_{t'}(\alpha)) = 4\epsilon_{t'}(\alpha).
\end{equation}
Now, given that $\alpha$ is deactivated whenever the deactivation rule is applied again, we will either collect (or counterfactually infer) one reward value for $\alpha$ after $t'$, for $t' < T$, or will not collect any reward value again, for $t'=T$. 
As a result, $\nu_{t'}(\alpha) \leq \nu_{T}(\alpha) \leq 1 + \nu_{t'}(\alpha)$ and
%
thus, using also Eq.~\ref{eq:delta-alpha}, we can conclude that, for any suboptimal $\alpha$, it holds that:
\begin{equation}\label{eq:delta-alpha-upper-bound}
    \Delta(\alpha) \leq O(\epsilon_{T}(\alpha)) = O\left( \sqrt{\frac{\log T}{\nu_{T}(\alpha)} }\right) = O\left( \sqrt{\frac{\log T}{\nu_{T}(\alpha)} }\right).
\end{equation}
From the above, it immediately follows that anytime we pull $\alpha$, we suffer average regret $\Delta(\alpha)$.
Consequently, for 
any
time $t \leq T$, the total average regret due to pulling arm $\alpha$, which we 
denote as $\EE[R(\alpha\,;\,t) \given \Ecal]$, is given by $\EE[R(\alpha\,;\,t) \given \Ecal] = n_{t}(\alpha) \cdot \Delta(\alpha)$, where $n_{t}(\alpha)$ denotes the number of times
that arm $\alpha$ has been pulled until $t$. Thus, for any time $t \leq T$, the total average regret conditioned on $\Ecal$ is given by:
\begin{equation}\label{eq:cumul-reward-upper-bound}
    \EE[R(t) \given \Ecal] = \sum_{\alpha \in \Acal} \EE[R(\alpha\,;\,t) \given \Ecal] = \sum_{\alpha \in \Acal} n_{t}(\alpha)\cdot\Delta(\alpha) \leq \sum_{\alpha \in \Acal} n_{t}(\alpha) \cdot O\left( \sqrt{\frac{\log T}{\nu_{t}(\alpha)} }\right).
\end{equation}
%
Now, we will derive a lower bound for $\nu_{t}(\alpha)$. Assume that, until time step $t \leq T$, the deactivation rule has been applied $n>0$ times. 
Let $m_1, m_2, ..., m_n$, where $0 \leq m_i \leq m, i=1,..,n$, be the number of arms that are not 
active after the $i$-th time the deactivation rule is applied. 
Before applying the deactivation rule, we collect (or counterfactually infer) one reward value for 
each arm with a number of pulls that is logarithmic to the number of active arms. 
Before applying the rule for the first time, all $m$ arms are active, so we will need up to $\lceil \log m \rceil$ rounds to collect a reward for each active arm. 
Then, given that $m_1$ arms are deactivated, $m-m_1$ remain active, so we will need up to $\lceil \log (m - m_1) \rceil$ rounds to collect a reward for each active arm until the deactivation rule is applied again. 
As a result, 
we have that:
\begin{equation*}
    t \leq \lceil \log m \rceil + \lceil \log (m - m_1) \rceil + \lceil \log (m-m_2) \rceil + \ldots + \lceil\log (m - m_{n-1}) \rceil.
\end{equation*}
Each logarithmic term in the above equation corresponds to the process of collecting one reward for each arm and, as a result, one reward for $\alpha$. 
As a result, the above has $\nu_{t}(\alpha)$ logarithmic terms and, given that $0 \leq m_i, i=1,..,n$, 
we have that:
\begin{equation*}
    t \leq \lceil \log m \rceil + \lceil \log (m - m_1) \rceil + \lceil \log (m-m_2) \rceil + \ldots
    + \lceil\log (m - m_{n-1}) \rceil  \leq \nu_{t}(\alpha) \cdot  \lceil \log m \rceil.
\end{equation*}
Therefore, we can conclude that $\nu_{t}(\alpha) \geq t/  \lceil \log m \rceil$
and,
using Eq.~\ref{eq:cumul-reward-upper-bound}, we have that:
\begin{align*}
    \EE[R(t) \given \Ecal] \leq \sum_{\alpha \in \Acal} n_{t}(\alpha) \cdot O\left( \sqrt{\frac{\log T}{\nu_{t}(\alpha)} }\right)&\leq \sum_{\alpha \in \Acal} n_{t}(\alpha) \cdot O\left( \sqrt{\frac{\log m \log T}{t}}\right)\\ & = O\left( \sqrt{\frac{\log m \log T}{t}}\right) \cdot \sum_{\alpha \in \Acal} n_{t}(\alpha).
\end{align*}
Further, since
it must hold that $\sum_{\alpha \in \Acal} n_{t}(\alpha) = t$ by definition, we can conclude that:
\begin{equation}\label{eq:regret-clean-event}
    \EE[R(t) \given \Ecal] \leq O\left( \sqrt{\frac{\log m \log T}{t}}\right) \cdot t = O\left( \sqrt{t \log m  \log T}\right).
\end{equation}

Let us now lift the assumption that $\Ecal$ holds. 
From Hoeffding'{}s bound and a union bound, it readily follows that:
\begin{equation}\label{eq:bad-event-prob-bound}
    \PP[\Ecal] \geq 1 - mT \delta \Rightarrow 1 - \PP[\Ecal] \leq m T \delta \Rightarrow \PP[\Bar{\Ecal}] \leq m T \delta,
\end{equation}
where $\delta = 2/T^4$. 
Moreover, given that the rewards take values in $\{0,1\}$, it holds that at time step $t$, $R(t) \leq t$. 
Consequently, combining Eqs.~\ref{eq:exp-regret-clean-event},~\ref{eq:regret-clean-event} and~\ref{eq:bad-event-prob-bound}, we can conclude that:
\begin{align*}
\EE[R(t)] &= \EE[R(t) \given \Ecal] \PP[\Ecal] + \EE[R(t) \given \Bar{\Ecal}]  \PP[\Bar{\Ecal}] \leq \EE[R(t) \given \Ecal] + \EE[R(t) \given \Bar{\Ecal}]  \PP[\Bar{\Ecal}]\\& = O\left(\sqrt{t \log m  \log T}\right) +  \frac{2tmT}{T^4},
\end{align*}
where the first inequality uses that $\PP[\Ecal] \leq 1$.
Finally, using that, by assumption, $\sqrt{m} \leq T$, the above becomes:
\begin{align*}
    \EE[R(t)] &\leq O\left(\sqrt{t \log m  \log T}\right) +  \frac{2tmT}{T^4} \\& \leq O\left(\sqrt{t \log m \log T}\right) +  \frac{2tT^3}{T^4} =  O\left(\sqrt{t \log m  \log T}\right) + O\left(\frac{t}{T}\right) \\ &= O\left(\sqrt{t \log m  \log T}\right).
\end{align*}
This concludes the proof.

\clearpage
\newpage

%


\section{Additional Details about the Human Subject Study Setup}
\label{app:experiment-details}

\begin{figure}
    \centering
    \includegraphics[width=.8\linewidth]{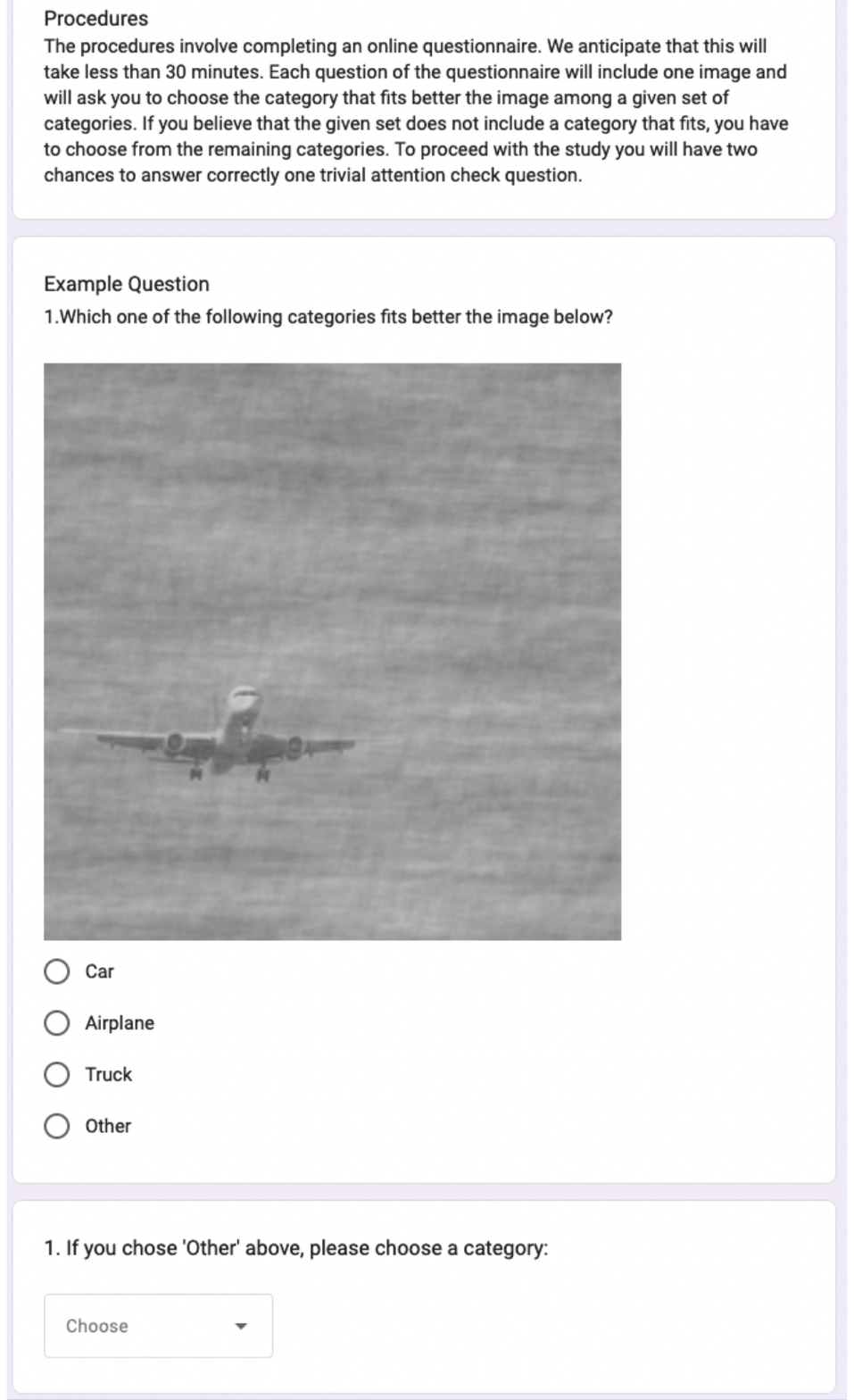}
    \caption{Procedures and example question included in the consent form that Prolific workers had to fill in order to participate in our study under the lenient implementation of our systems.}
    \label{fig:procedures-eg-q-lenient}
\end{figure}

\xhdr{Human subject study consent form} Figure~\ref{fig:consent-a} shows screenshots of the consent form that Prolific workers had to fill in order to participate in our study under the strict implementation of our system.  
We used a similar consent form for our study under the lenient implementation of our system except for the ``Procedures'' 
and ``Example Question'' sections, which we show in Figure~\ref{fig:procedures-eg-q-lenient}. 

\xhdr{Dataset}
The ImageNet16H dataset~\citep{steyvers2022bayesian} was created using $1{,}200$ unique images labeled into $207$ different~fine-grained categories from the ImageNet Large Scale Visual Recognition Challenge (ILSRVR) 2012 dataset~\citep{russakovsky2015imagenet}.
%
%
More specifically, in the ImageNet16H dataset, 
each of the above images was used to generate four noisy images, each with a different amount of phase noise distorsion $\omega \in \{80, 95, 110, 125\}$,
and 
each of the above fine-grained categories was mapped into one out of $16$ coarse-grained categories (\ie, chair, oven, knife, bottle, keyboard, clock, boat, bicycle, airplane, truck, car, elephant, bear, dog, cat, and bird), which serve as ground truth labels\footnote{Note that, by mapping the fine-grained categories used in the ILSRVR 2012 dataset into coarse-grained categories, one essentially gets rid of any potential label disagreement among annotators in the ILSRVR 2012 dataset.}.
%
%
%
%
The amount of phase noise distortion controls the difficulty of the classification task; the higher the noise, the more difficult the classification task. 
In our experiments, we used all $1{,}200$ noisy images with $\omega=110$ because, under such noise value, humans sometimes, but not always, succeed at solving the prediction task (\ie, the empirical success probability achieved by the human experts on their own was $0.760$).   

\xhdr{Implementation details}
We implemented our algorithms on Python 3.10.9 using the following libraries:
\begin{itemize}
    \item NumPy 1.24.1 (BSD-3-Clause License).
    \item Pandas 1.5.3 (BSD-3-Clause License).
    \item Scikit-learn 1.2.2 (BSD License).
\end{itemize}
For reproducibility, we used a fixed random seed in all random procedures (a different one) for each realization of the algorithms. 
Similarly, we used a fixed random seed to randomly pick the $120$ images of the calibration set.

\xhdr{Stratifying images and users}
We stratify the images into groups of similar difficulty, where we measure the difficulty of each image using the empirical success probability of experts predicting its ground truth label. 
More specifically, for each image we first compute the empirical success probability over all experts'{} predictions. Then, we stratify the images into $5$ mutually exclusive groups as follows:
\begin{enumerate}[label=(\roman*),leftmargin=0.8cm,noitemsep,nolistsep]
    \item[---] Highest difficulty: images with empirical success probability within the $20\%$ percentile of the empirical success probabilities of all images. 
    \item[---] Medium to high difficulty: images with empirical success probability within the $40\%$ percentile and outside the $20\%$ percentile of the empirical success probabilities of all images.
     \item[---] Medium difficulty: images with empirical success probability within the $60\%$ percentile and outside the $40\%$ percentile the of the empirical success probabilities of all images. 
     \item[---] Low difficulty: images with empirical success probability within the $80\%$ percentile and outside the $60\%$ percentile the of the empirical success probabilities of all images. 
     \item[---] Lowest difficulty: images with empirical success probability outside the $80\%$ percentile of the empirical success probabilities of all images. 
\end{enumerate}
%
%
We follow a similar method to stratify experts into two groups based on their level of competence, where we measure the level of competence of an expert using the empirical success probability of predicting the ground truth label across all the predictions that she made. As a result we consider the $50\%$ of users with the highest empirical success probability as the experts with high level of competence and the rest $50\%$ as experts with low level of competence.

\newpage
\clearpage
\begin{figure}
    \centering    
    \subfloat[Purpose of the study and procedures]{
    \includegraphics[width=0.6\linewidth, angle=-90]{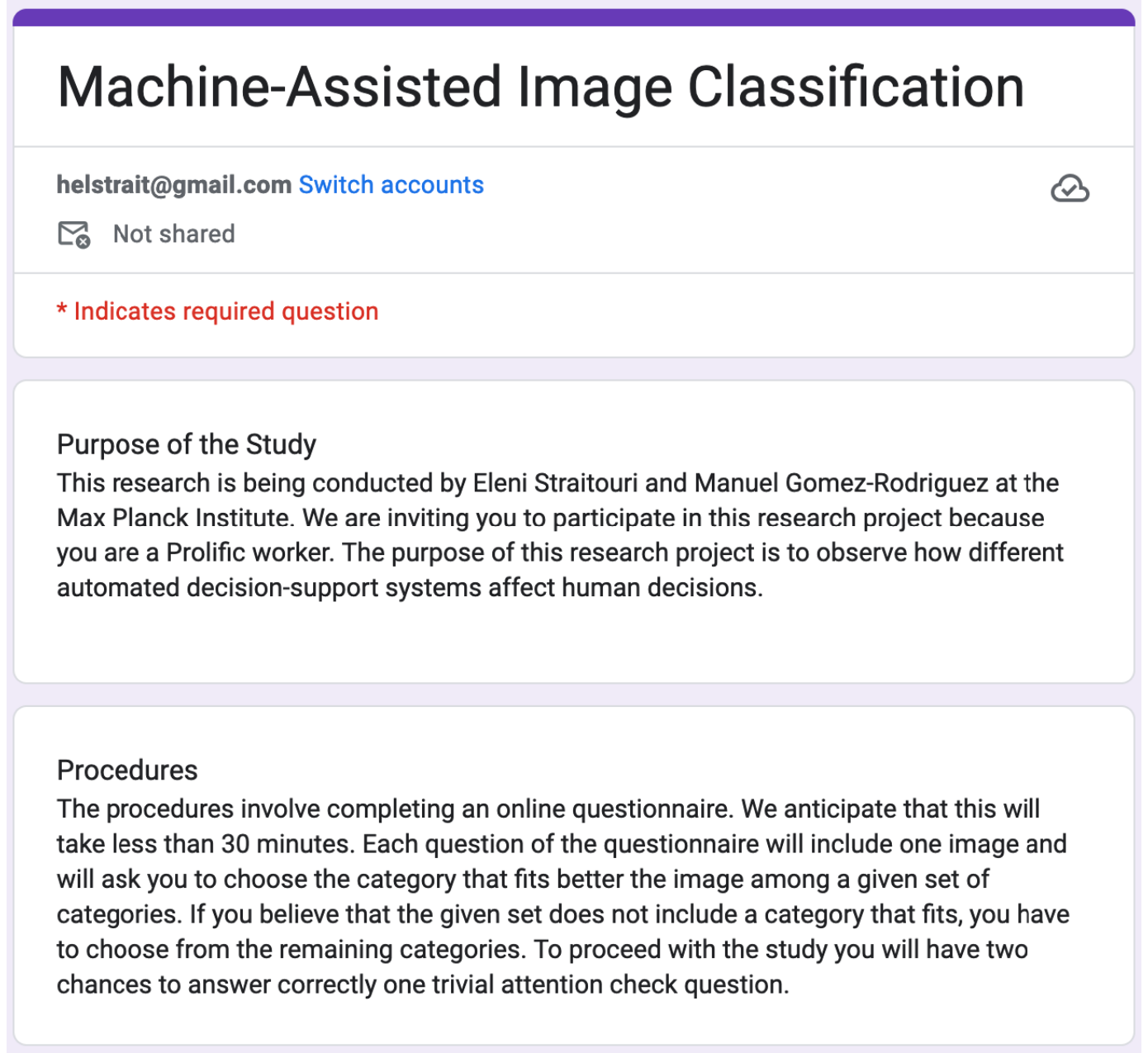}
    }
    \caption{The consent form including a detailed description of the study processes that Prolific workers had to read and fill in order to participate in our human subject study. The procedures describe use of the decision support systems under the strict implementation. The consent form continues in Figures~\ref{fig:consent-b} and~\ref{fig:consent-c}.}
    \label{fig:consent-a}
\end{figure}
\newpage
\clearpage
\begin{figure}
    \ContinuedFloat
    \centering
    \subfloat[Example question, potential risk and discomforts and potential benefits]{
    \includegraphics[width=0.8\linewidth]{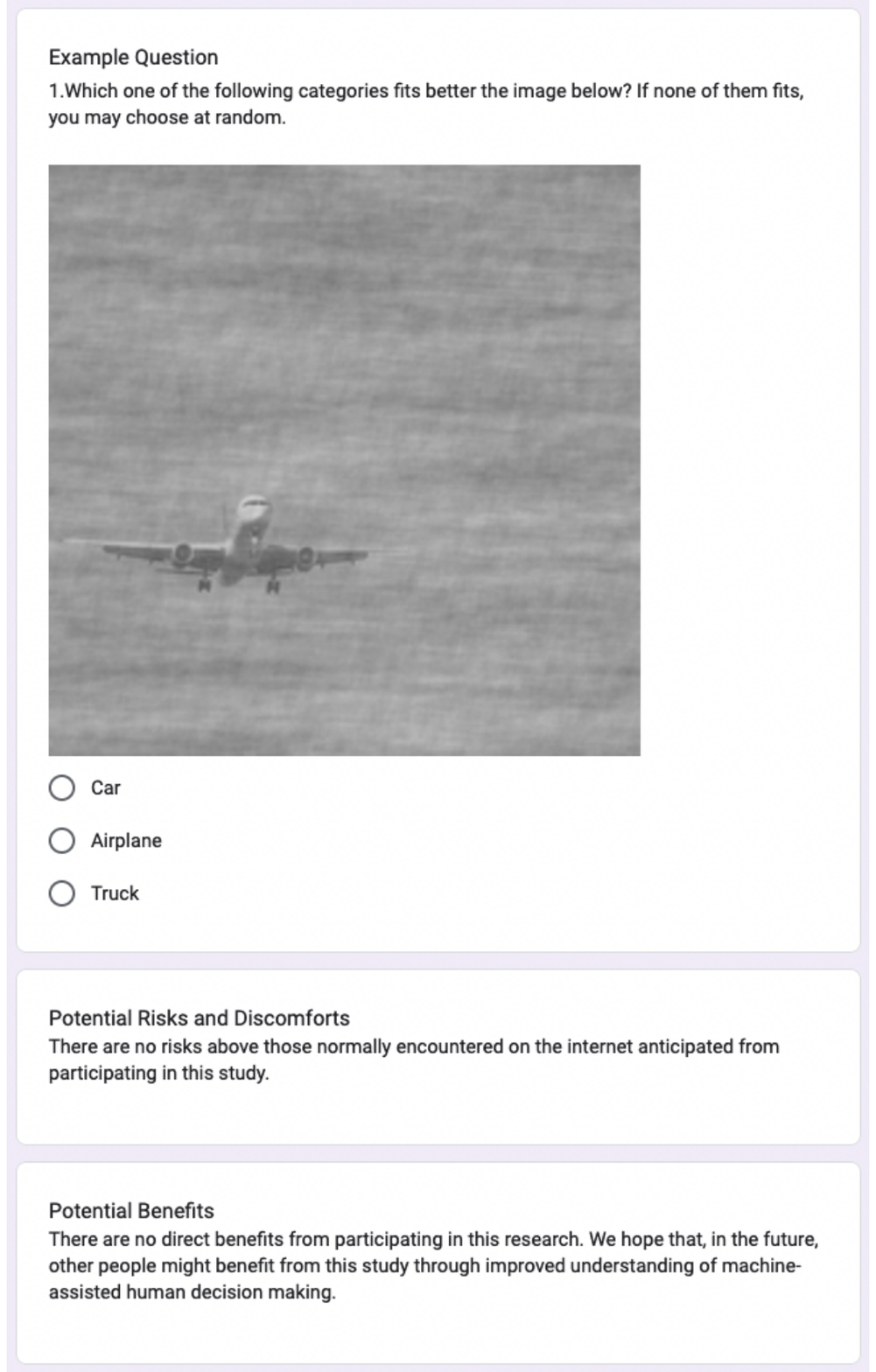}
    \label{fig:consent-b}
    }
    \caption{Consent form continued.}
\end{figure}
\newpage
\clearpage
\begin{figure}
    \ContinuedFloat
    \centering
    \subfloat[Confidentiality, compensation, right to withdraw, questions, and consent]{\includegraphics[width=0.8\linewidth]{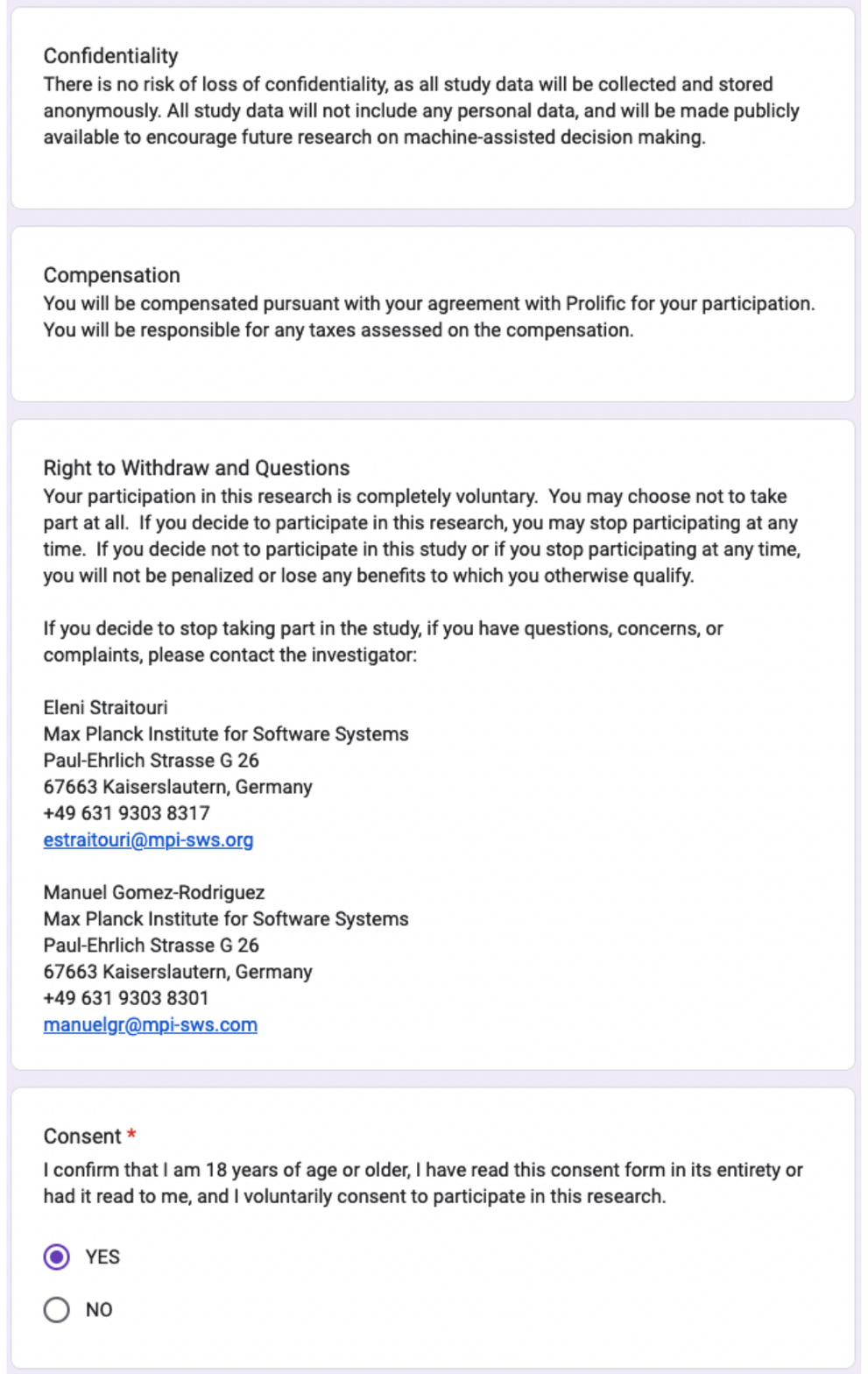}
    \label{fig:consent-c}
    }
    \caption{Consent form continued.}
\end{figure}
\newpage
\clearpage
\newpage
\clearpage

\section{Expert Success Probability vs. Prediction Set Size} 
\label{app:monotonicity}
\begin{figure}[h]
    \centering
    \subfloat[All experts]
    {
    \begin{tabular}{@{}ccc@{}}
    \includegraphics[width=.3\linewidth]{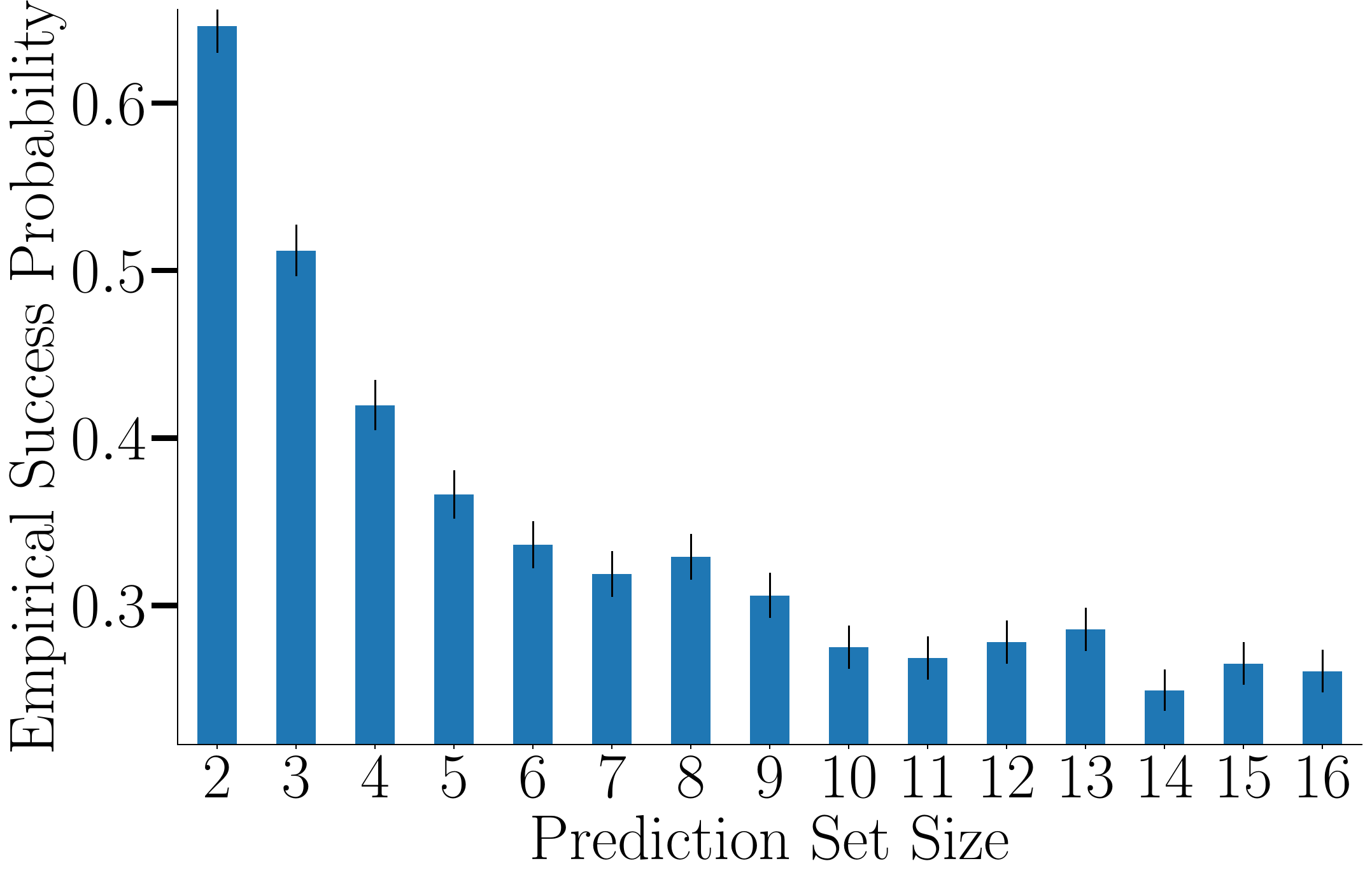}
    &
    \includegraphics[width=.3\linewidth]{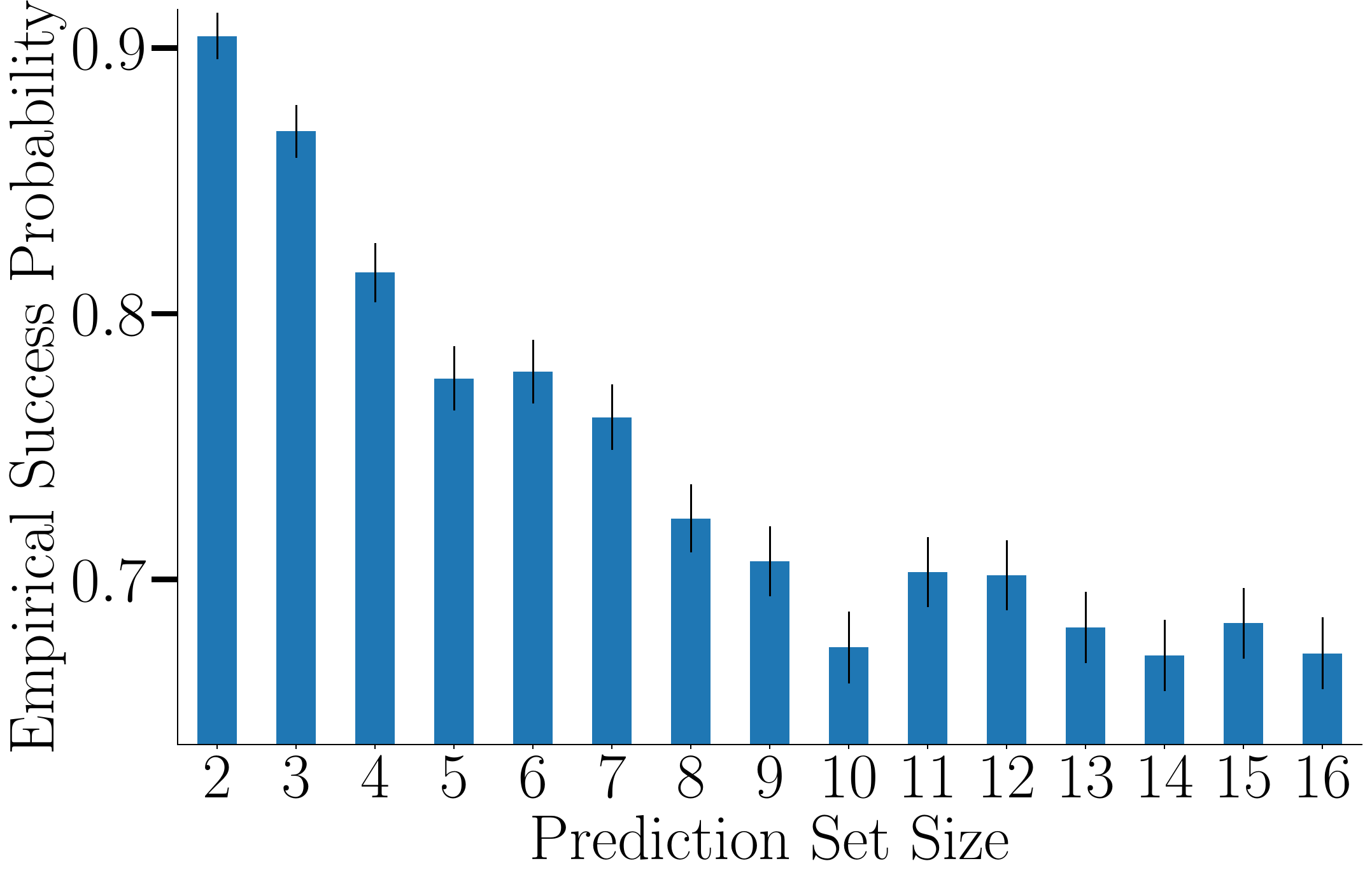}
    & 
    \includegraphics[width=.3\linewidth]{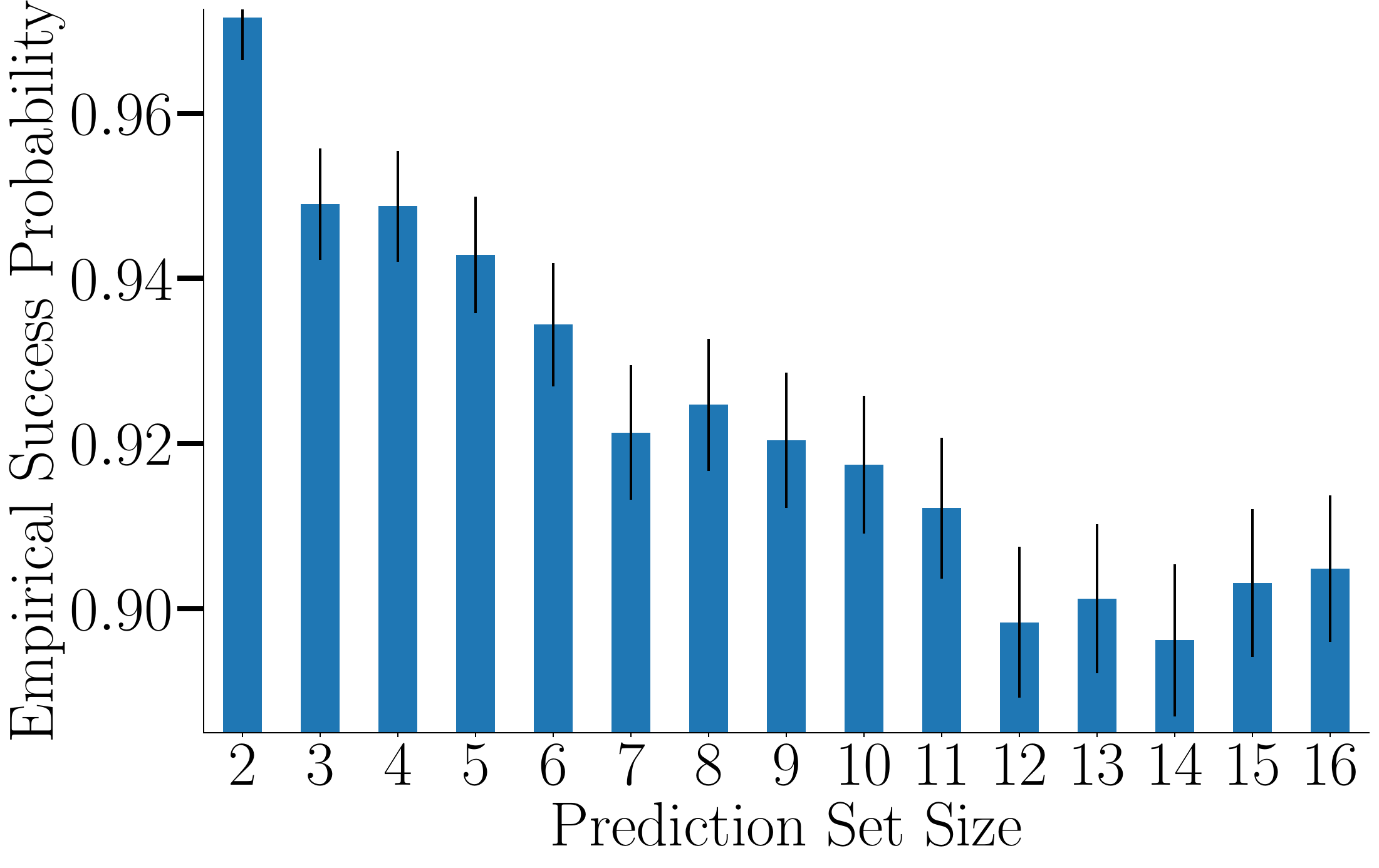}
    \\
    \small High difficulty & \small Medium to high difficulty & \small Medium difficulty \\
    \end{tabular}
    } \\
    \subfloat[Experts with low level of competence]
    {
    \begin{tabular}{@{}ccc@{}}
    \includegraphics[width=.3\linewidth]{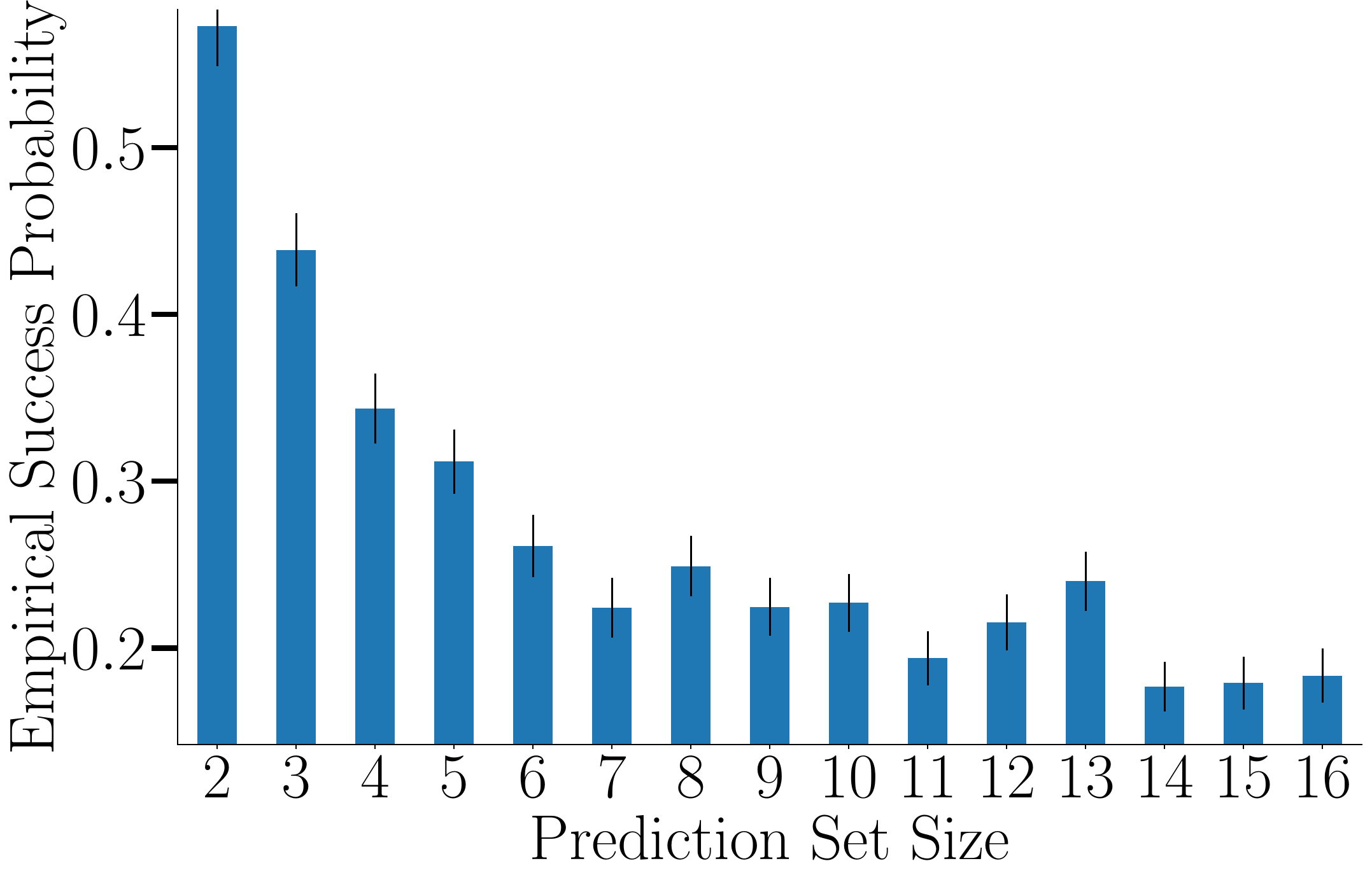}
    &
    \includegraphics[width=.3\linewidth]{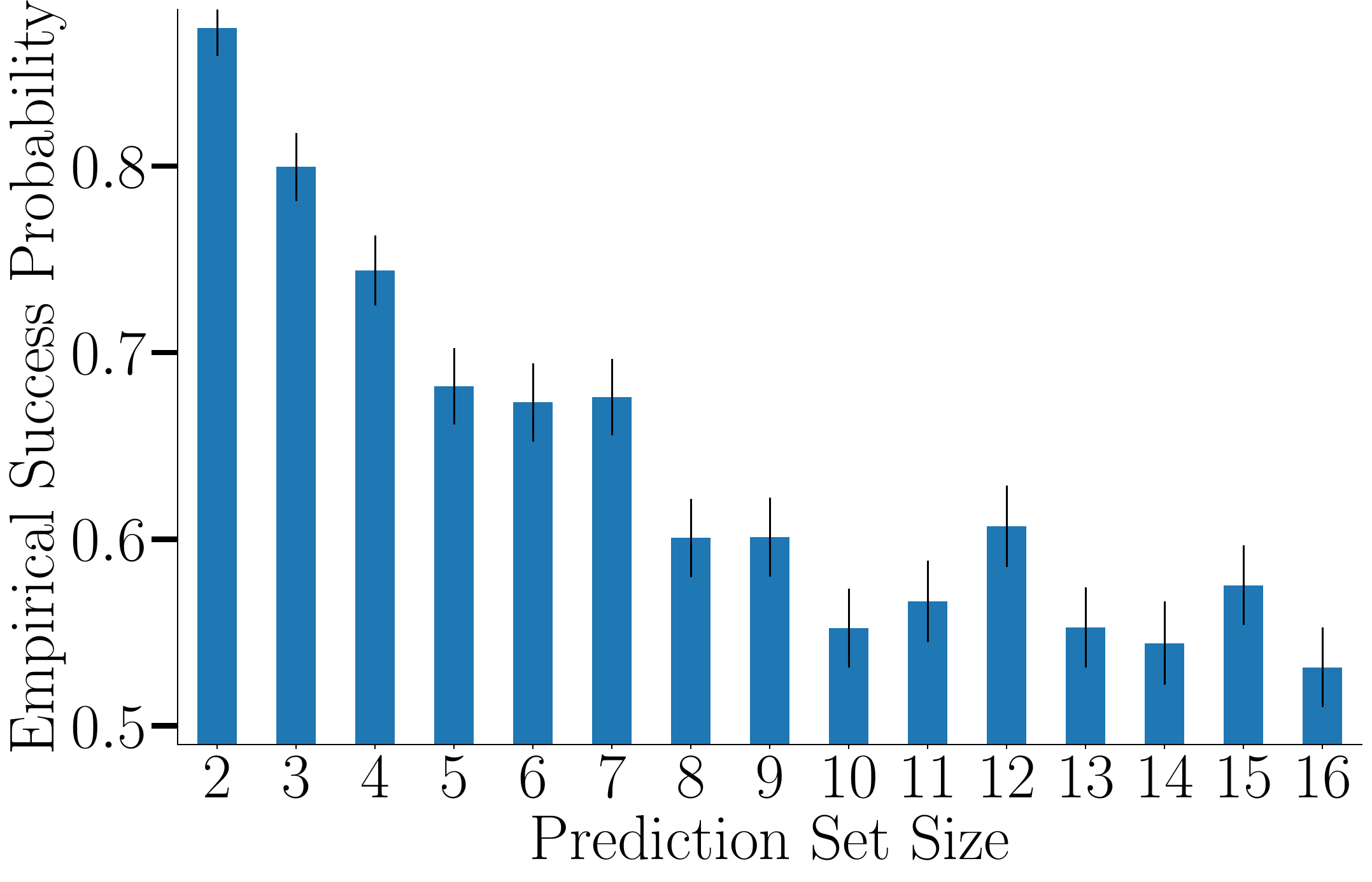}
    & 
    \includegraphics[width=.3\linewidth]{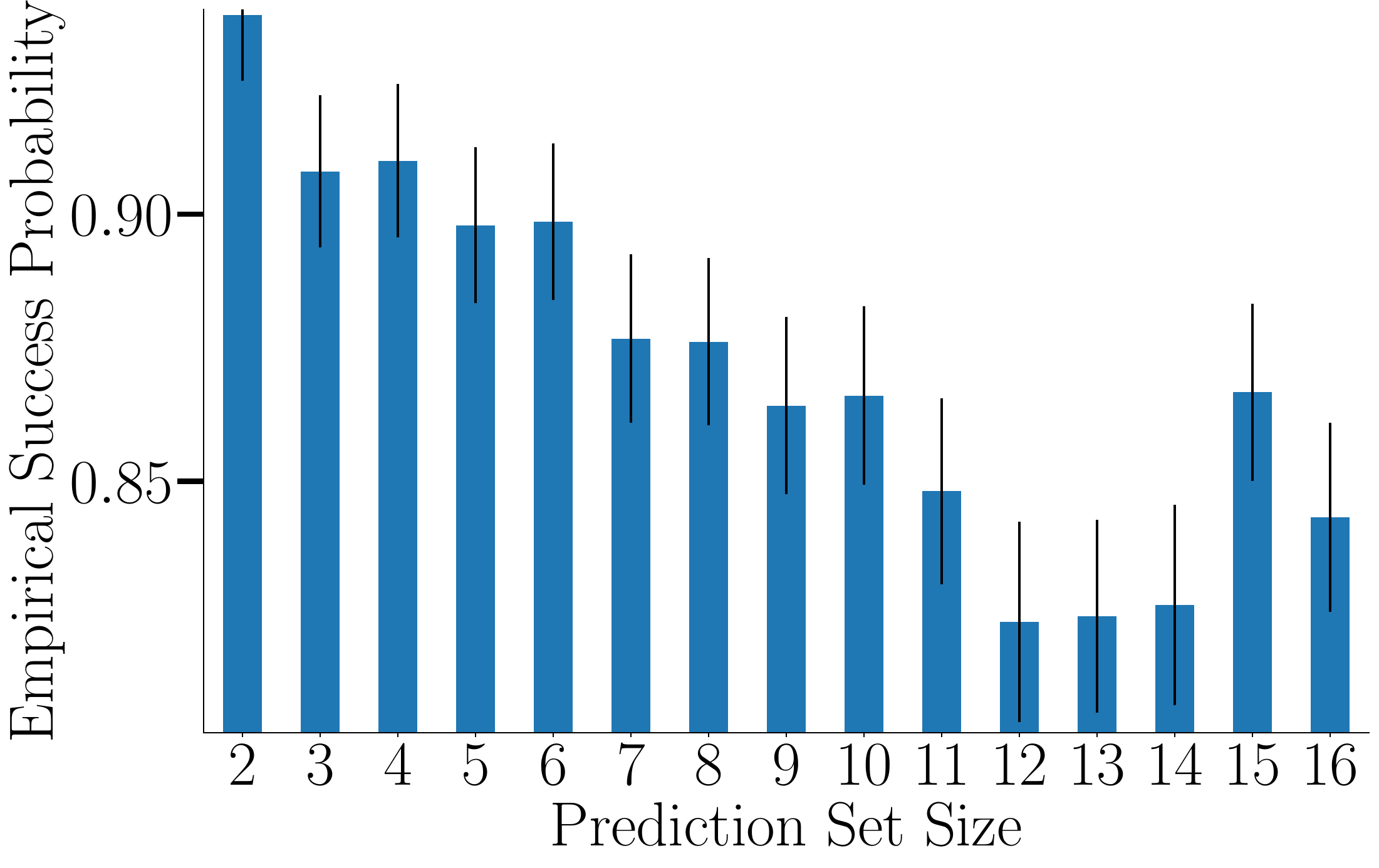}
    \\
    \small High difficulty & \small Medium to high difficulty & \small Medium difficulty \\
    \end{tabular}
    } \\
    \subfloat[Experts with high level of competence]
    {
    \begin{tabular}{@{}ccc@{}}
    \includegraphics[width=.3\linewidth]{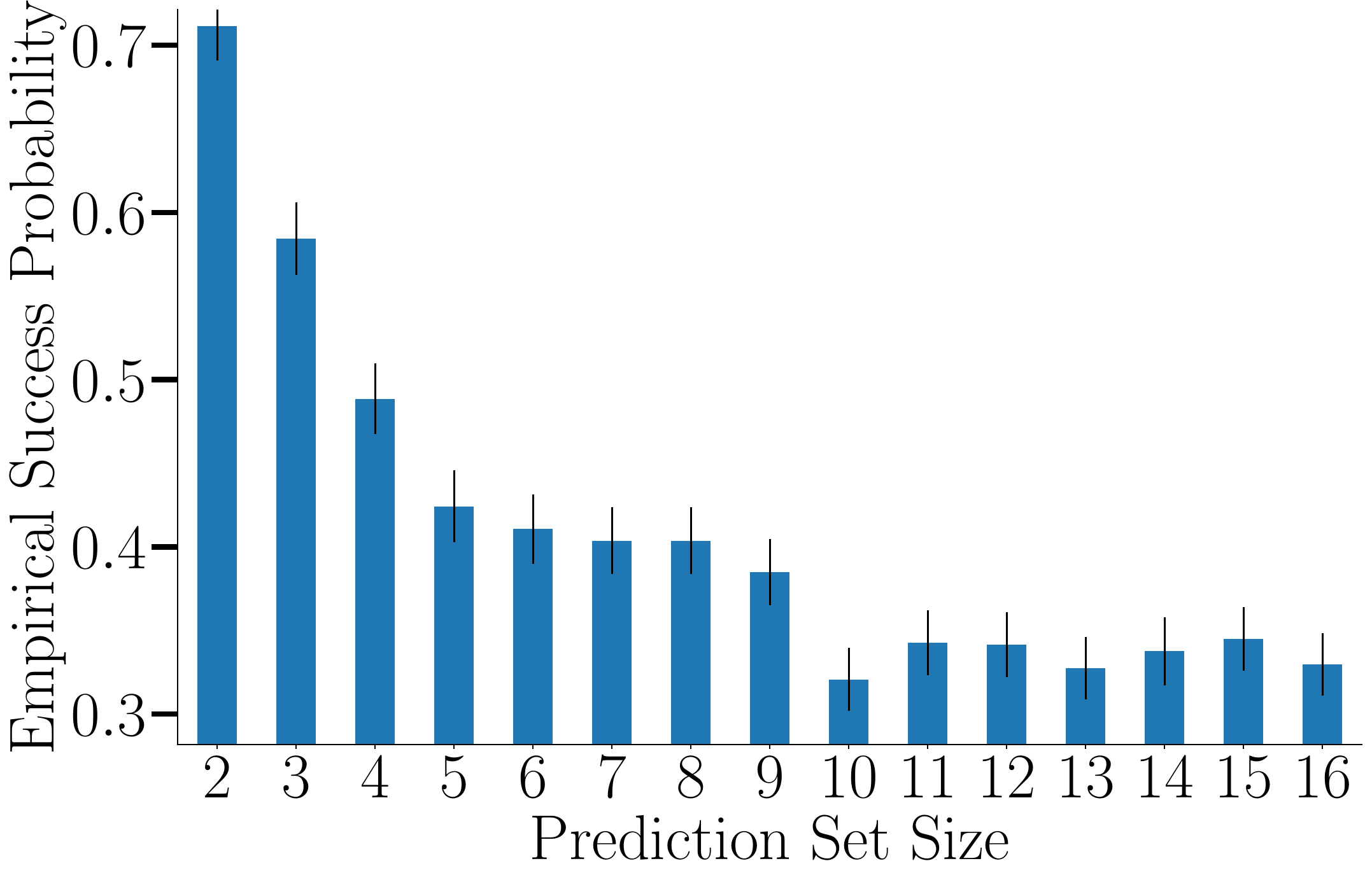}
    &
    \includegraphics[width=.3\linewidth]{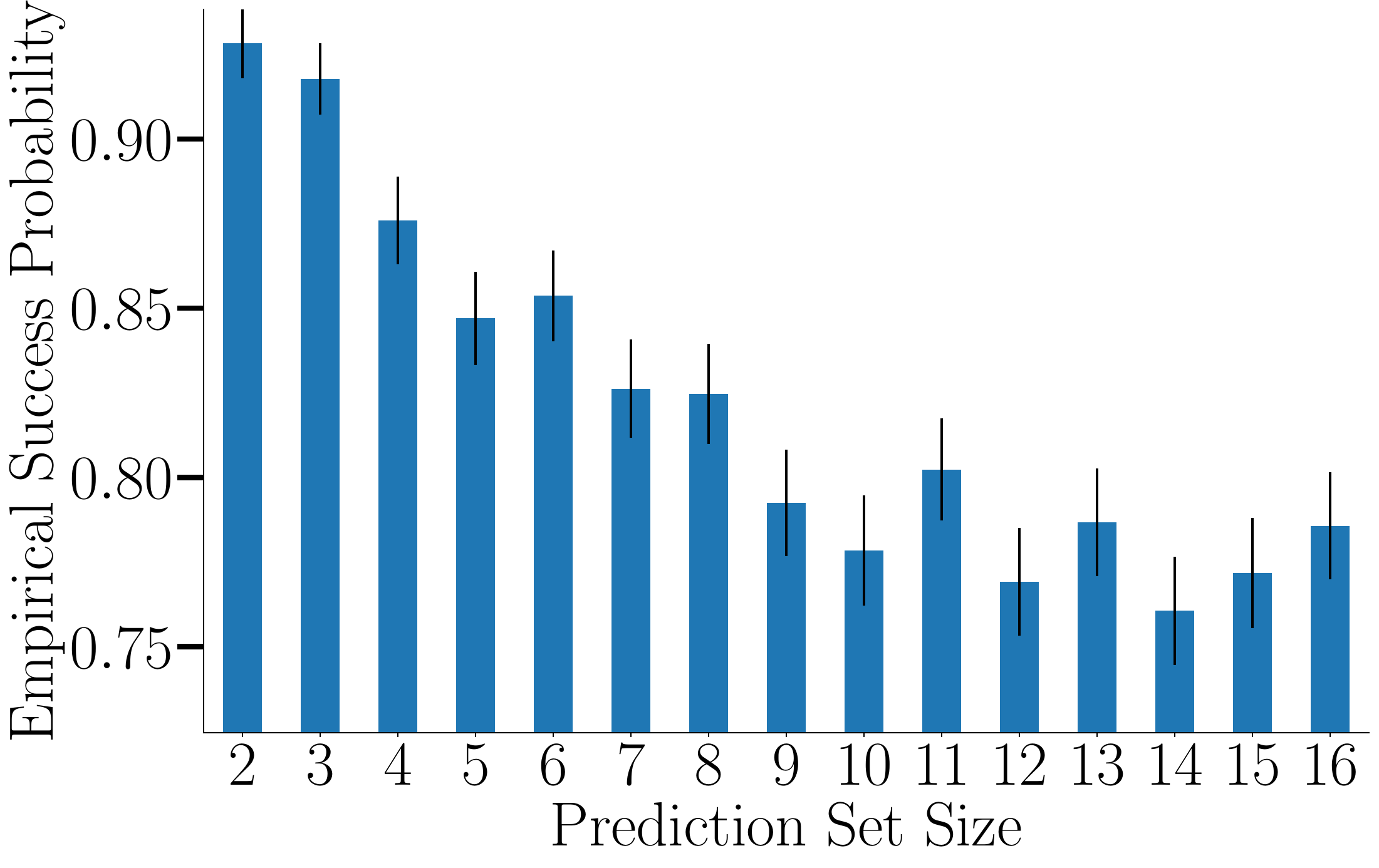}
    & 
    \includegraphics[width=.3\linewidth]{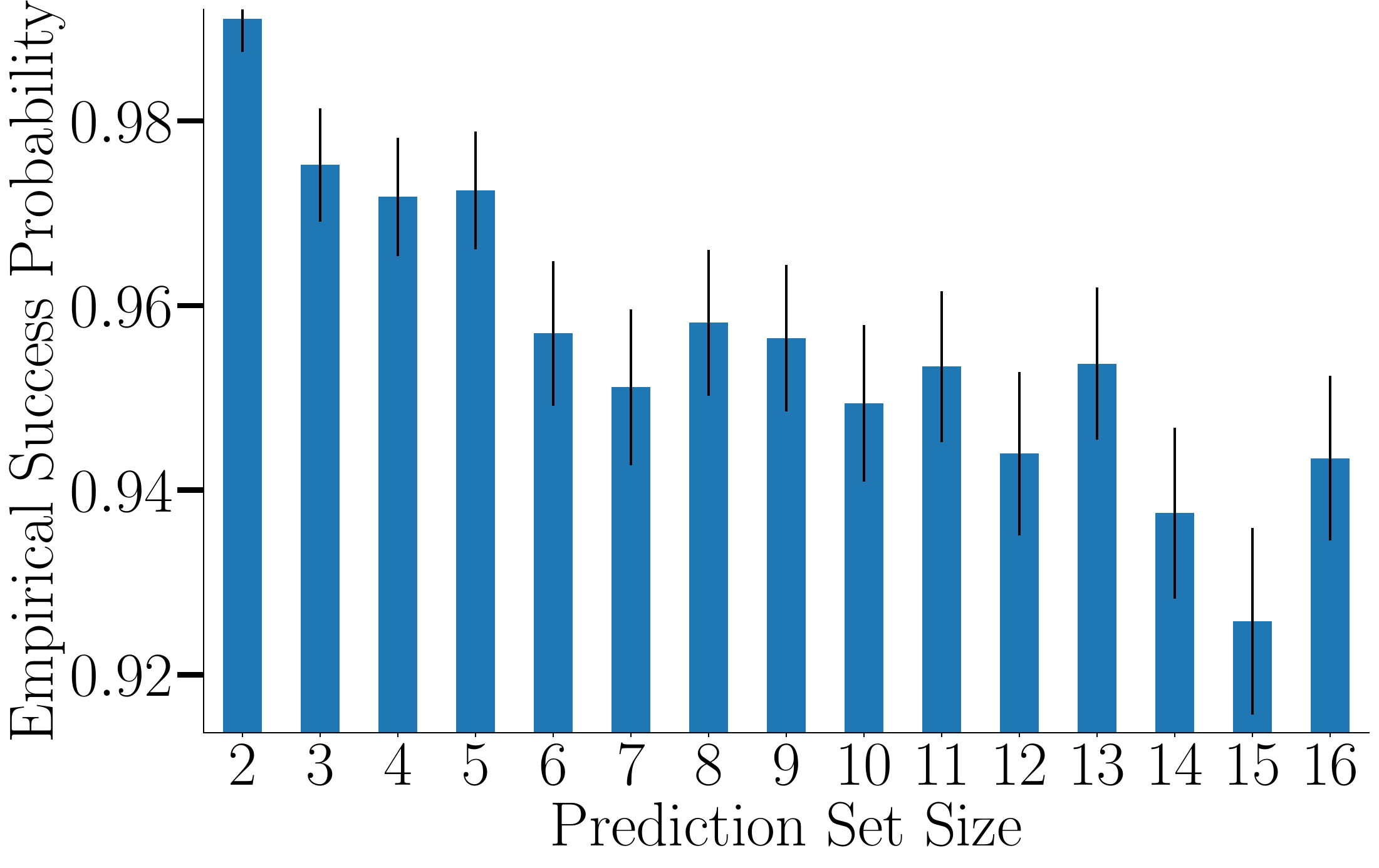}
    \\
    \small High difficulty & \small Medium to high difficulty & \small Medium difficulty \\
    \end{tabular}
    }
    \caption{Empirical success probability per prediction set size averaged across (a) all experts, (b) experts with low level of competence, 
    and (c) experts with high level of competence, for images of high difficulty, medium to high difficulty and medium difficulty. 
    In all panels, we have only considered prediction sets that included the true label and thus have omitted showing the empirical success probability for singletons, as it is always $1$. 
    Error bars denote standard error.
    }
    \label{fig:acc-set-size-across-experts}
\end{figure}

Figure~\ref{fig:acc-set-size-across-experts} shows the empirical success probability per prediction set size across images with different difficulty levels and experts with different levels of competence.
%
%
%

\newpage
\clearpage

\section{Sensitivity Analysis to Violations of the Counterfactual Monotonicity Assumption}
\label{app:sensitivity-montonicity}
In this section, we study the sensitivity of counterfactual SE and counterfactual \texttt{UCB1} to violations of the counterfactual monotonicity assumption. 
In what follows, 
we first describe how we post-process the experts'{} predictions gathered in our human subject study to artificially increase the amount of counterfactual monotonicity violations 
and then discuss the performance of counterfactual SE and counterfactual \texttt{UCB1} under different amounts of counterfactual monotonicity violations.
Throughout the section, we focus on the experts'{} predictions using the strict implementation
of our system.
\begin{figure}[ht]
    \centering
    \subfloat[High difficulty]
    {
    \begin{tabular}{@{}ccc@{}}
    \includegraphics[width=.3\linewidth]{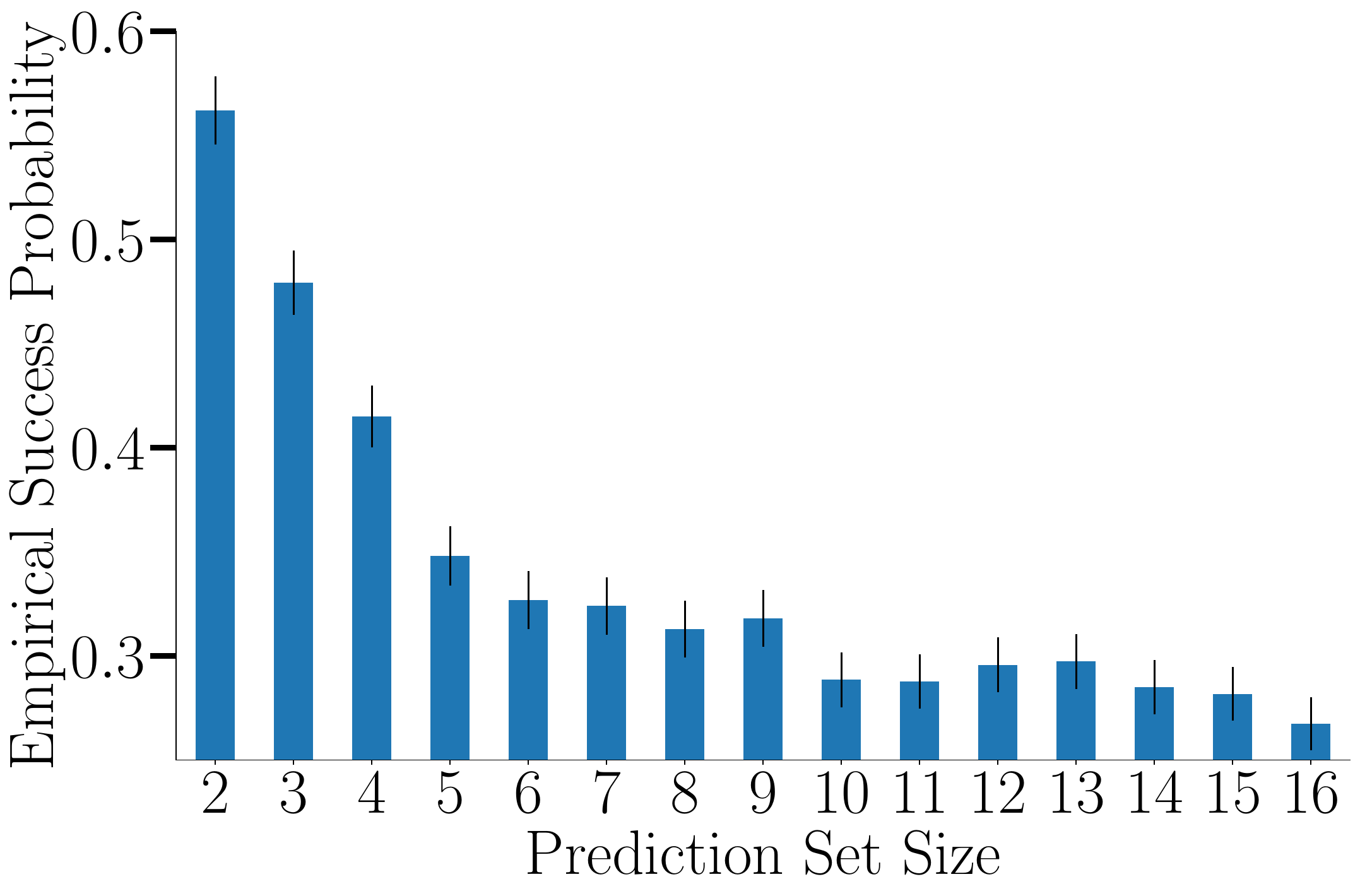}
    &
    \includegraphics[width=.3\linewidth]{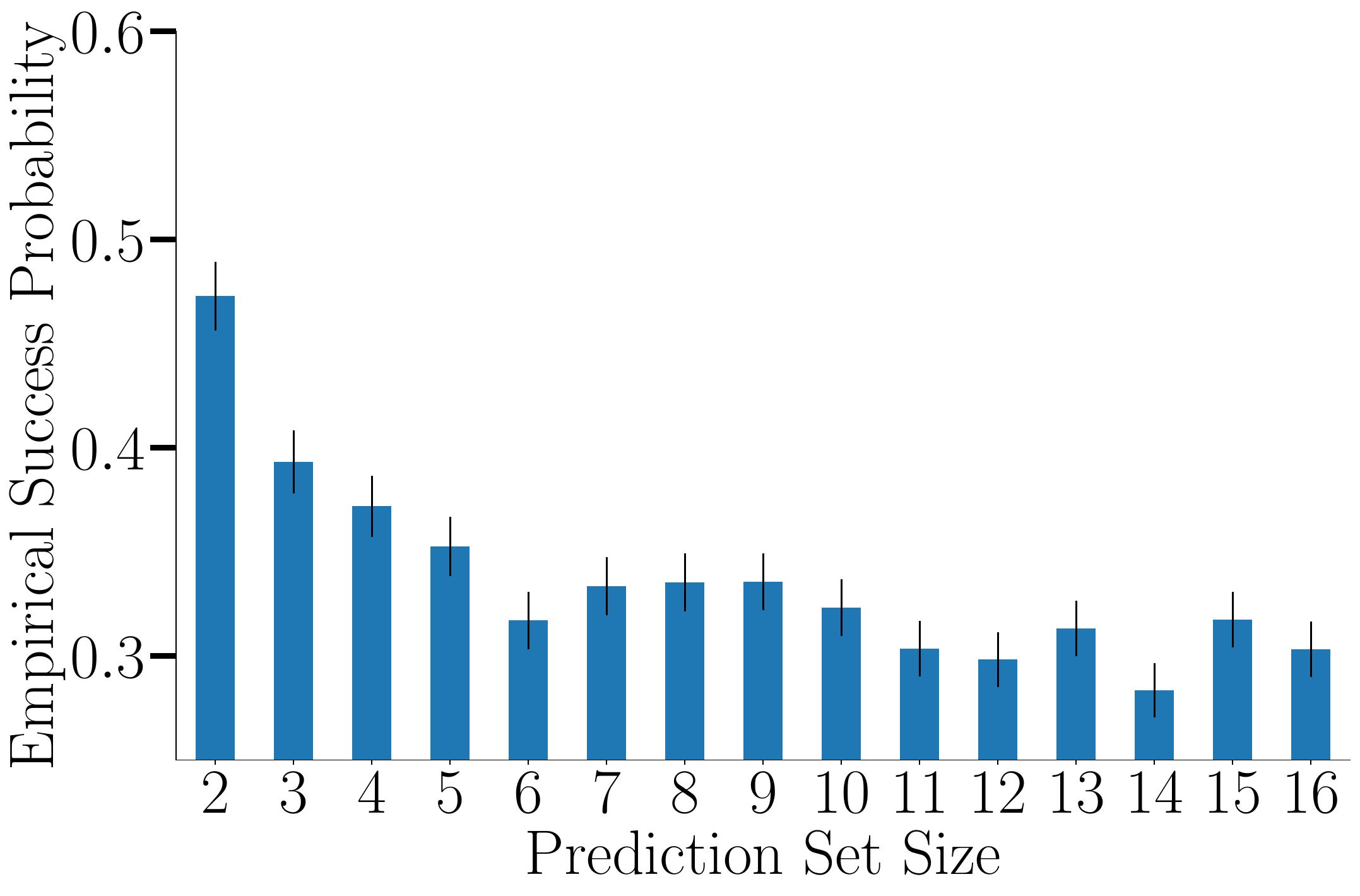}
    & 
    \includegraphics[width=.3\linewidth]{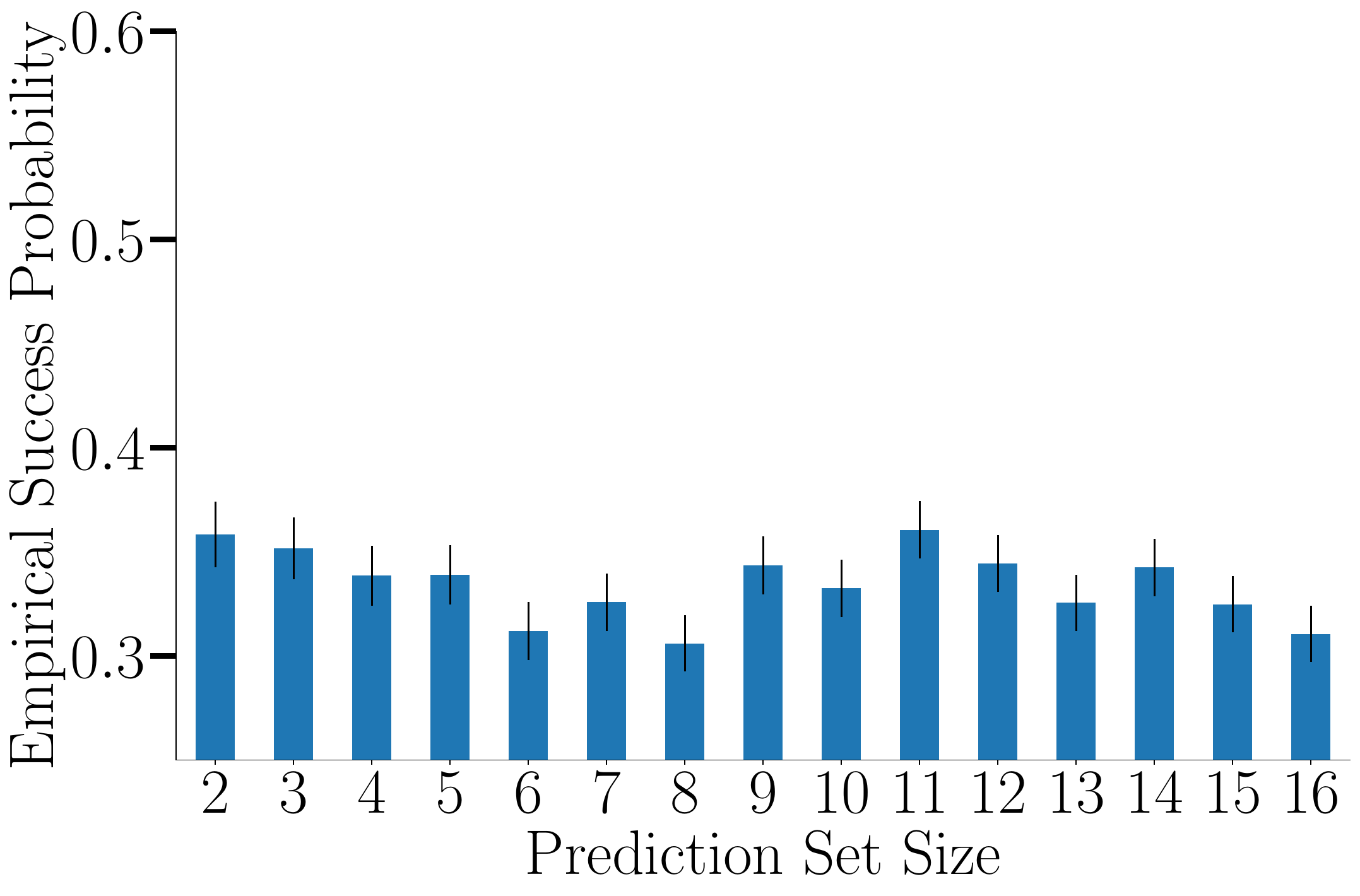}
    \\
    \small \qquad $p_v=0.3$ & \small  \qquad $p_v=0.6$ & \small  \qquad $p_v=1.0$ \\
    \end{tabular}
    } \\
    \subfloat[Medium to high difficulty]
    {
    \begin{tabular}{@{}ccc@{}}
    \includegraphics[width=.3\linewidth]{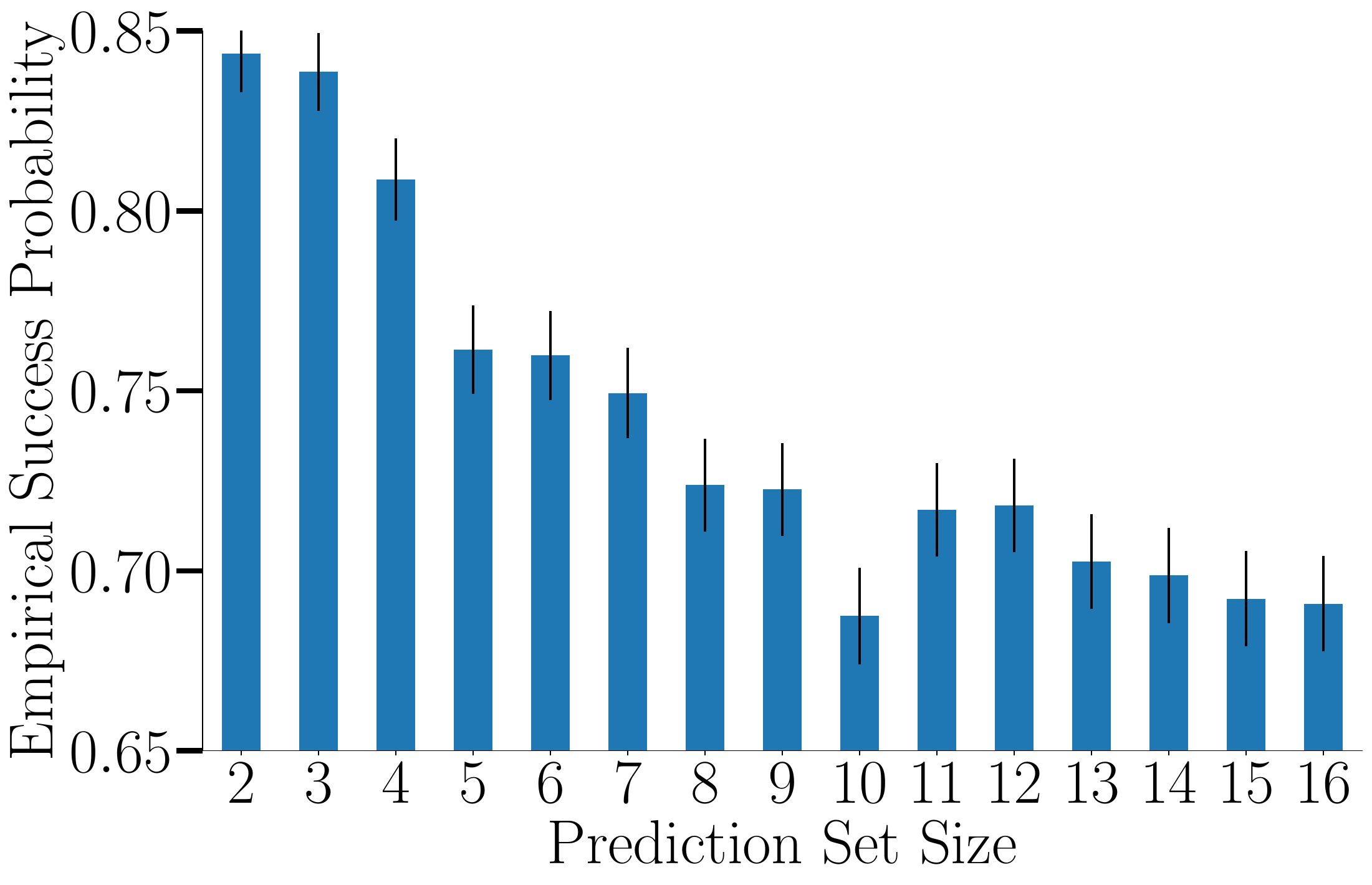}
    &
    \includegraphics[width=.3\linewidth]{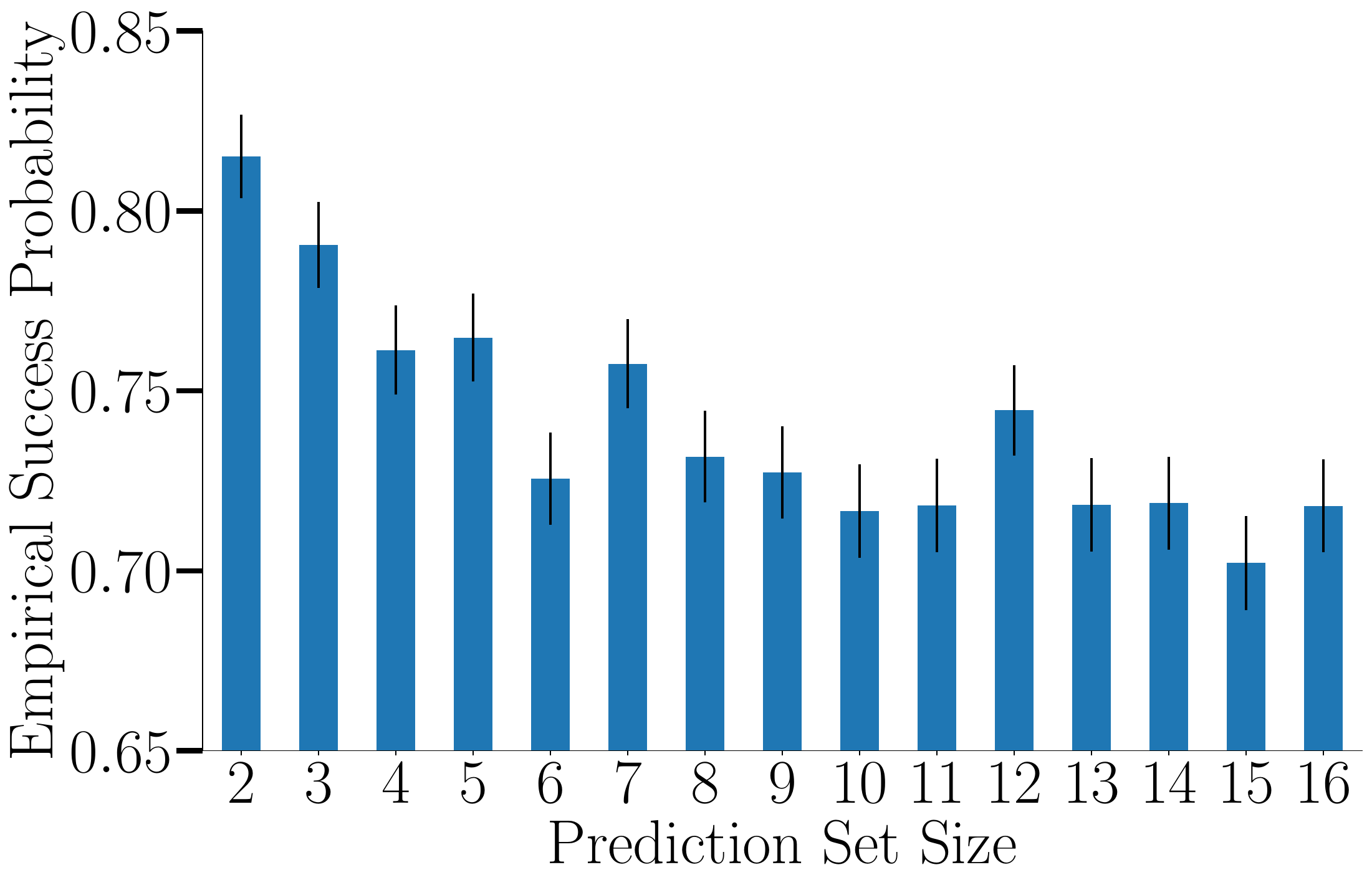}
    & 
    \includegraphics[width=.3\linewidth]{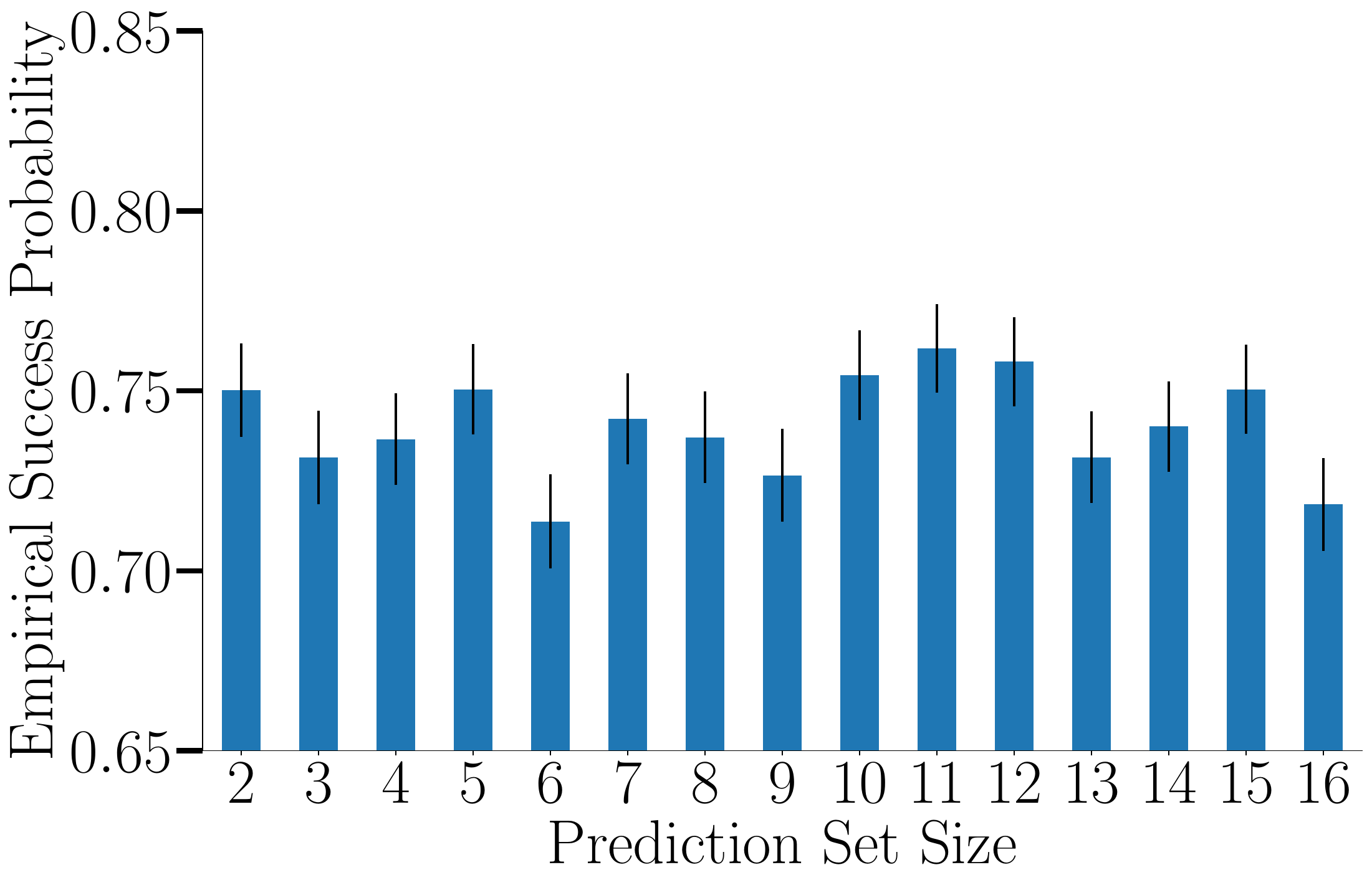}
    \\
    \small \qquad $p_v=0.3$ & \small \qquad $p_v=0.6$ & \small  \qquad $p_v=1.0$ \\
    \end{tabular}
    } \\\subfloat[Medium difficulty]
    {
    \begin{tabular}{@{}ccc@{}}
    \includegraphics[width=.3\linewidth]{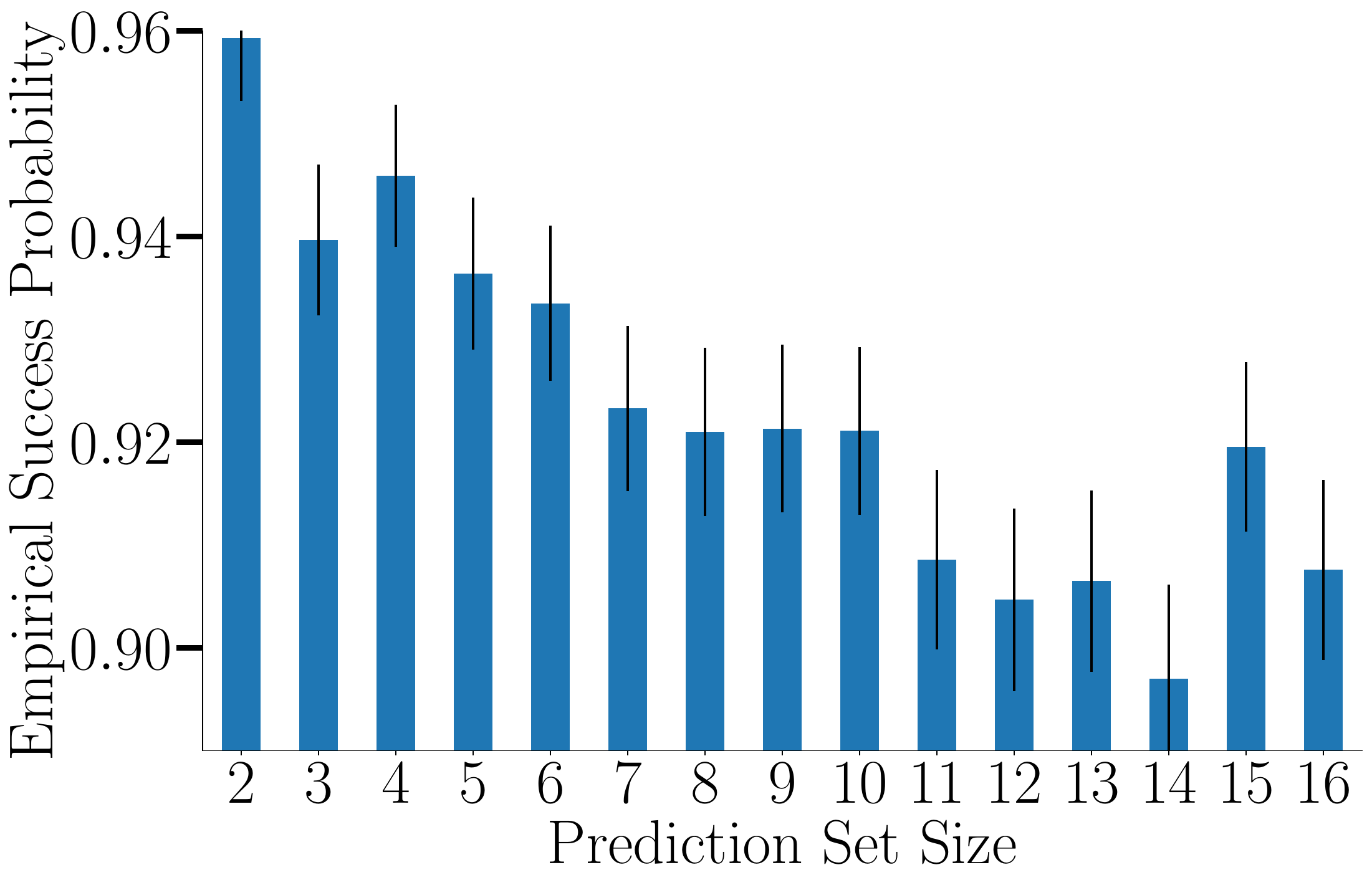}
    &
    \includegraphics[width=.3\linewidth]{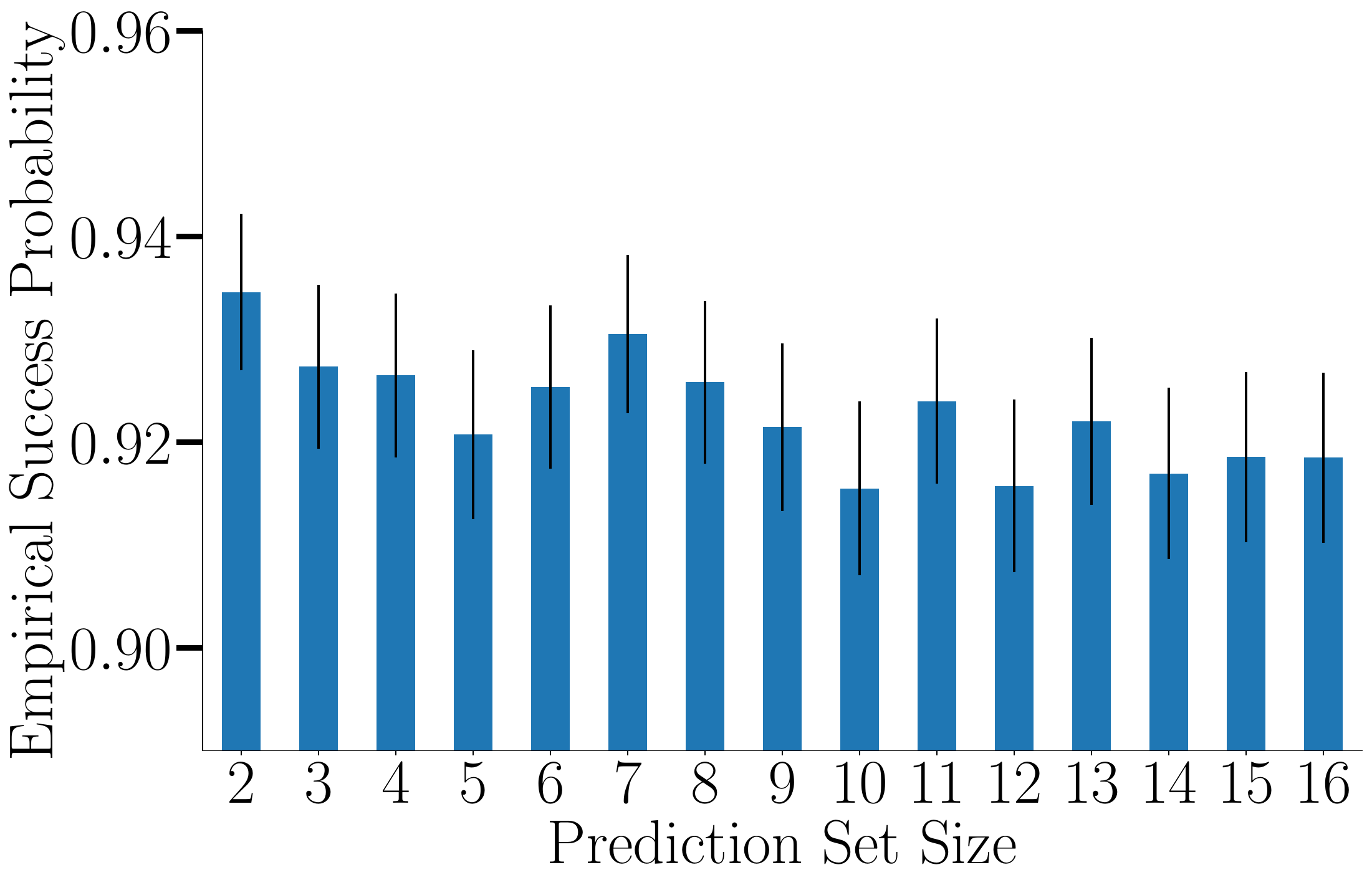}
    & 
    \includegraphics[width=.3\linewidth]{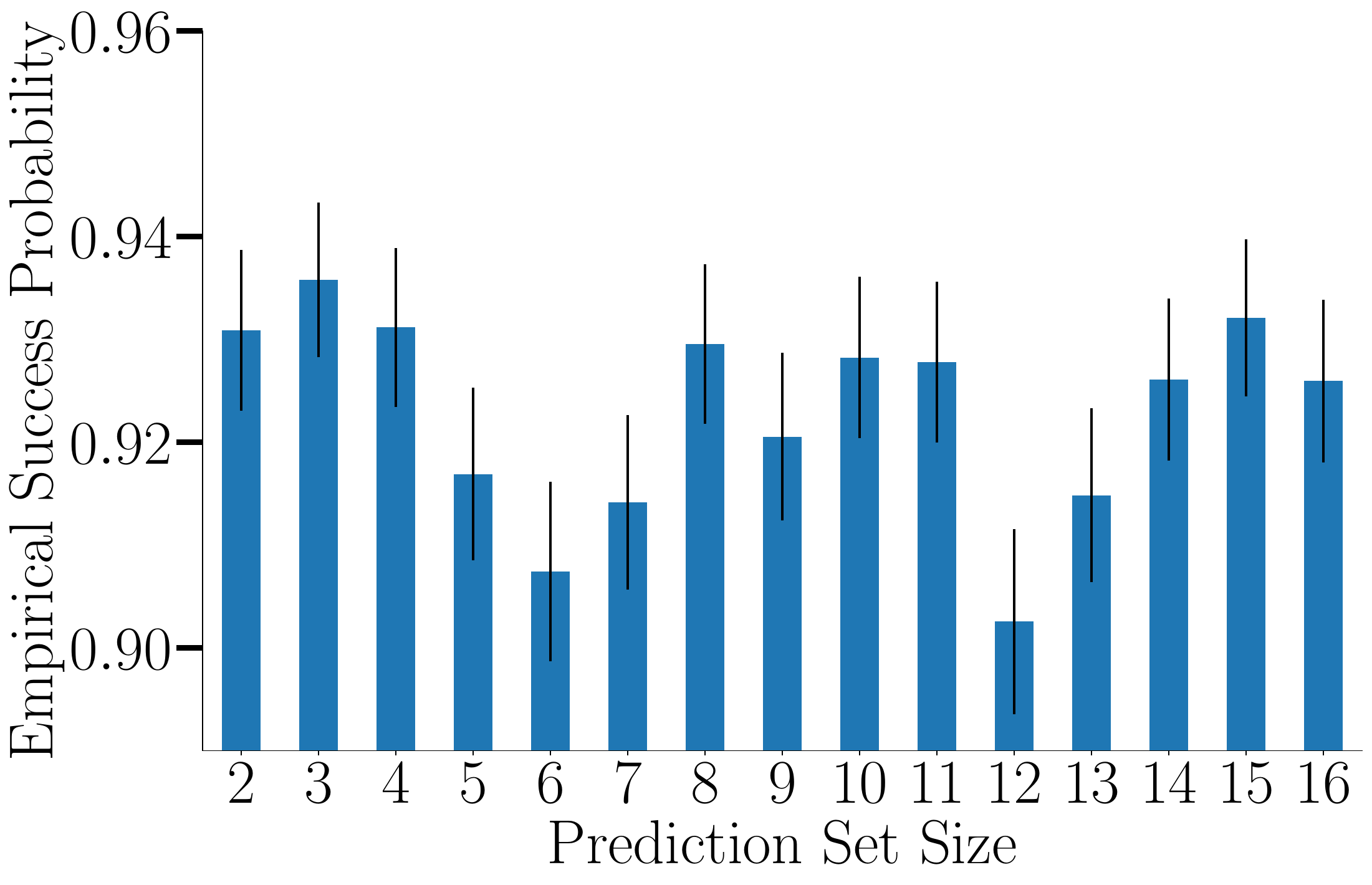}
    \\
    \small \qquad $p_v=0.3$ & \small  \qquad  $p_v=0.6$  & \small \qquad $p_v = 1.0$  \\
    \end{tabular}
    } 
    \caption{Empirical success probability per prediction set size averaged across all experts for images with (a) the highest difficulty, (b) medium to high difficulty, and (c) medium difficulty under different amounts of counterfactual monotonicity violations controlled by $p_v$. 
    In all panels, we have only considered prediction sets that included the true label and thus have omitted showing the empirical success probability for singletons, as it is always 1. 
    Error bars denote standard error.}
    \label{fig:acc-set-size-violations}
    \vspace{-3mm}
\end{figure}

\xhdr{Experimental setup} 
%
Since counterfactual monotonicity lies within level three in the ``ladder of causation''~\citep{pearl2009causal}, 
we cannot directly increase (nor estimate!) the amount of counterfactual monotonicity violations in the experts' predictions gathered in our human subject study.
However, we can increase the amount of interventional monotonicity violations 
(\ie, how frequently Eq.~\ref{asm:success-probability-monotonicity} is violated) and,
since interventional monotonicity is a necessary condition for counterfactual monotonicity, 
we argue that we are indirectly increasing the amount of counterfactual monotonicity violations.
To this end, we randomly select a fraction $p_v$ of the images used in our human subject study and, 
for each of these images, 
we randomly permute the values $\II\{\hat{y}_{\Ccal_{\alpha}}=y\}$ across
pairs of experts' predictions $\hat{y}_{\Ccal_{\alpha}}$ and prediction sets $\Ccal_{\alpha}(x)$ 
such that $y \in \Ccal_{\alpha}(x)$.
%
%
%
%
Here, the larger the fraction $p_v$ of images we permute their labels, the larger the 
amount of interventional (counterfactual) monotonicity violations, as shown in 
Figure~\ref{fig:acc-set-size-violations}.
In the Figure, we stratify the images into groups of similar difficulty, following the
procedure described in Appendix~\ref{app:experiment-details}, to demonstrate that the
effect of the above random permutations is more apparent in the images of higher 
difficulty.\footnote{We do not show the empirical success probability for images of low 
difficulty because it is originally almost always $1$ and thus it is not affected by the 
random permutations.}
%
%
%
%

\begin{figure}[ht]
    \centering
    \subfloat[$p_v = 0.3$ ]{
    \includegraphics[width=.3\linewidth]{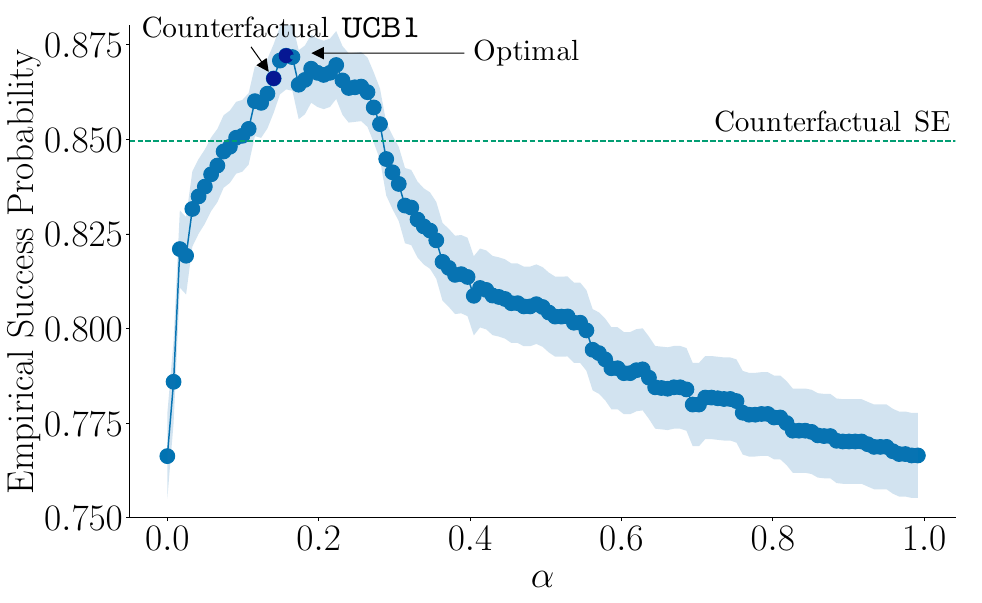}
    }
    \subfloat[$p_v=0.6$]{
    \includegraphics[width=.3\linewidth]{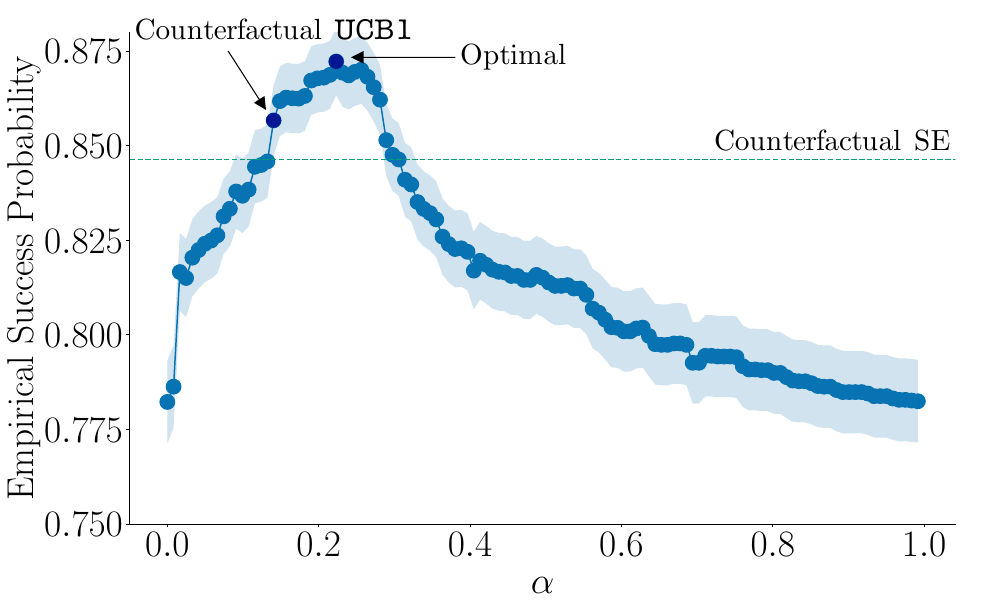}
    }
    \subfloat[$p_v=1.0$ ]{
    \includegraphics[width=.3\linewidth]{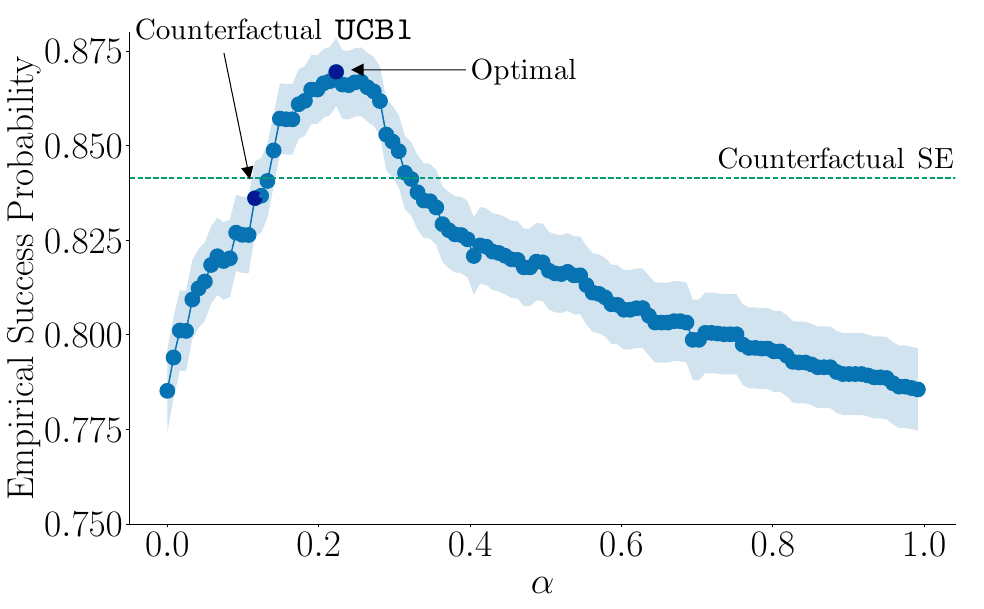}
    }
    \caption{Empirical success probability achieved by all experts across all images 
    using our system $\Ccal_{\alpha}$ with different $\alpha$ values under different
    amounts of counterfactual monotonicity violations controlled by $p_v$.
    In each panel, we annotate the optimal $\alpha^{*}$, the $\alpha$ value found by counterfactual \texttt{UCB1}, as well as the average success probability achieved by the set of $\alpha$ values that remain active after running counterfactual
    SE. 
    The average accuracy of the classifier used by our system is $0.848$ and the empirical success probability achieved by the experts on their own is between $0.766$ and $0.785$.
    Here, note that, since we also permute experts' predictions whenever the prediction set is $\Ycal$, the empirical success probability achieved by the experts on their own changes
    across panels.
    The shaded area corresponds to a 95\% confidence interval.}
    \label{fig:acc-vs-alpha-strict-violations}
\end{figure}

\xhdr{Results} 
%
%
Figure~\ref{fig:acc-vs-alpha-strict-violations} shows (i) the empirical success probability
achieved by all experts using our system $\Ccal_{\alpha}$ against the full range of values 
of $\alpha \in [0, 1]$,
(ii) the optimal $\alpha^{*}$, 
(iii) the $\alpha$ value found by counterfactual \texttt{UCB1} 
and 
(iv) the average success probability achieved by the set of $\alpha$ values that
remain active after running counterfactual SE,
under different amounts of counterfactual monotonicity violations.
As expected, we find that the greater the amount of counterfactual monotonicity~vio\-la\-tions, the greater the difference between the empirical success probability achieved by our system with the optimal $\alpha^{*}$ and by our system with the $\alpha$ values found by counterfactual SE and counterfactual \texttt{UCB1}.
However, we also find that the performance degrades gracefully with respect to the 
amount of counterfactual monotonicity violations.
For example, even if we introduce monotonicity violations in the experts' predictions for all images, the empirical success probability achieved by all experts under the $\alpha$ value found by counterfactual \texttt{UCB1} is only $3.8$\% lower than under the optimal $\alpha^{*}$.
%
%
%
%
%

\newpage
\clearpage

\section{Expert Success Probability under the Strict and Lenient Implementation of our Systems}
\label{app:acc-vs-alpha}
\begin{figure}[!h]
    \centering    
    \includegraphics[width=.6\linewidth]{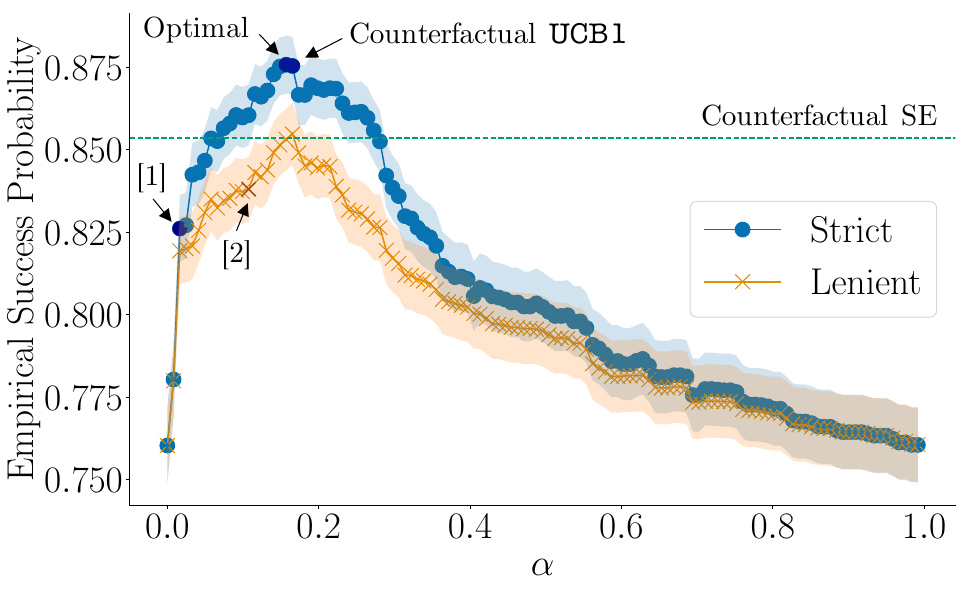}
    \caption{Empirical success probability achieved by all experts across all images using the strict and lenient implementation of our system with different $\alpha$ values. The annotated $\alpha$ values as well as the horizontal dashed line are the same as in Figure~\ref{fig:acc-vs-alpha}. The shaded areas correspond to a $95\%$ confidence interval.}
    \label{fig:acc-vs-alpha-full}
\end{figure}

Figure~\ref{fig:acc-vs-alpha-full} shows the empirical success probability under the full range of values of $\alpha \in [0,1]$. The results show that a strict implementation of our 
system consistently offers greater performance than a lenient implementation 
across the full spectrum of competitive $\alpha$ values. 
However, the results also show that, as the $\alpha$ value increases, the empirical success probability under both the strict and the lenient implementation of our system converges to the success probability of the experts'{} choosing on their own, \ie, choosing from $\Ycal$. 
This happens because, the larger the $\alpha$ value, the more often happens that the prediction set is the empty set
and thus we allow the expert to choose from $\Ycal$ under both implementations, as discussed in Footnote~7 in the
main paper.

\begin{figure}[ht]
    \centering 
    \quad
    \includegraphics[width=.64\linewidth]{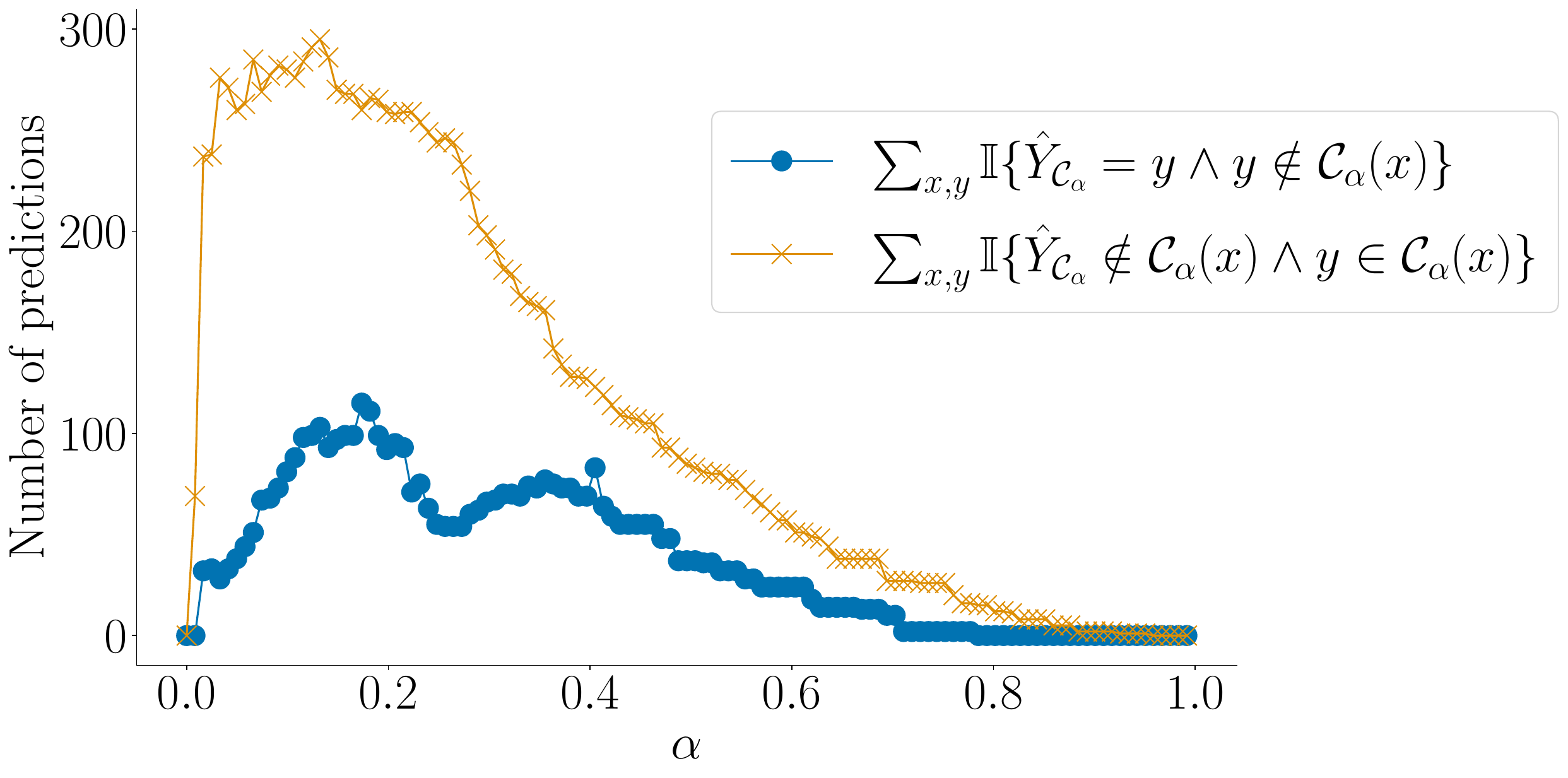}
    \caption{Number of experts'{} predictions in which, under the lenient implementation, the prediction sets do not contain the true label and the experts succeed (blue dots) and the prediction sets contain the true label and the experts fail (yellow crosses). 
    The sums are over the $1{,}080$ images not used in the calibration set.}
    \label{fig:n-predictions-disadvantage}
\end{figure}

Figure~\ref{fig:n-predictions-disadvantage} shows that the lenient implementation compares unfavorably against the strict implementation because, under the lenient implementation, the number of predictions in which the prediction sets do not contain the true label and the experts succeed
%
is consistently smaller than
%
%
the number of predictions in which the prediction sets contain
the true label and the experts fail.

\end{document}